\documentclass{article}

\usepackage{microtype}
\usepackage{graphicx}
\usepackage{subcaption}
\usepackage{booktabs}
\usepackage{enumitem}
\usepackage{hyperref}

\usepackage{icml2026}
\makeatletter
\icmlshowauthorstrue
\makeatother

\usepackage{amsmath}
\usepackage{amssymb}
\usepackage{mathtools}
\usepackage{amsthm}
\usepackage{placeins}
\usepackage[capitalize,noabbrev]{cleveref}
\usepackage{multirow}
\usepackage{xcolor}
\usepackage{colortbl}
\usepackage{subcaption}

\definecolor{oursrow}{RGB}{235,245,255}

\theoremstyle{plain}
\newtheorem{theorem}{Theorem}[section]

\theoremstyle{definition}
\newtheorem{definition}[theorem]{Definition}

\theoremstyle{remark}

\icmltitlerunning{BUMP: Brief Unexpected Mid-Interpolation Peak}

\makeatletter
\renewcommand{\Notice@String}{This paper was carried out in parallel with the work of Beatrix Benkő~\cite{benko2026quadratic}.}
\makeatother

\begin{document}

\twocolumn[
  \icmltitle{BUMP: Brief Unexpected Mid-Interpolation Peak}

  \icmlsetsymbol{equal}{*}

  \begin{icmlauthorlist}
    \icmlauthor{Melinda Flóra Kiss}{renyi,elte}
    \icmlauthor{Márton Muntag}{renyi}
    \icmlauthor{Adrián Csiszárik}{renyi}
    \icmlauthor{Dániel Varga}{renyi}
  \end{icmlauthorlist}

  \icmlaffiliation{renyi}{HUN-REN Alfréd Rényi Institute of Mathematics, Budapest, Hungary}
  \icmlaffiliation{elte}{Department of Computer Science, Eötvös Loránd University, Budapest, Hungary}

  \icmlcorrespondingauthor{Melinda Flóra Kiss}{melindafkiss@student.elte.hu}
  \icmlcorrespondingauthor{Adrián Csiszárik}{csadrian@renyi.hu}

  \icmlkeywords{deep learning, linear mode connectivity, model merging, model fusion, model ensembles}

  \vskip 0.3in
]

\printAffiliationsAndNotice{}  

\begin{abstract}
We identify an intriguing phenomenon when linearly interpolating between permutation-aligned neural networks: test accuracy briefly improves near the interpolation endpoints before dropping toward the midpoint, which we term BUMP (Brief Unexpected Mid-interpolation Peak). We observe that BUMP appears consistently across datasets, architectures, training regimes, and remains robust under substantial perturbations such as early checkpoints and extreme pruning. Moreover, controlled ablations suggest that it cannot be attributed to trivial regularization effects. While similar near-endpoint gains are visible in prior interpolation plots, they have not been explicitly documented or systematically and thoroughly analyzed. Leveraging this effect, we propose a simple, novel ensembling scheme: weight–prediction mixed ensembling that combines weight mixing with prediction-space mixing. Across our experiments, the resulting ensembling method consistently outperforms standard logit ensembling in test accuracy and yields improved calibration trade-off, highlighting complementary benefits of weight- and prediction-space mixing.
\end{abstract}

\section{Introduction}

Linear mode connectivity (LMC) has emerged as a useful lens to understand neural network loss landscapes \citep{frankle2020linear, entezari2022permutation}. As observed by \citet{ainsworth2022rebasin}, when the weights of independently trained, heavily overparameterized neural networks are aligned by optimizing over neuron permutations, the resulting networks become linearly mode connected. The standard way to quantify mode connectivity is the loss barrier, defined as the maximal increase in loss observed when linearly interpolating between two models. It is empirically found to occur near the midpoint of the interpolation loss curve, plotting the loss as a function of the $\lambda$ interpolation coefficient.

In this paper, we focus on another point along the interpolation curve. Our main observation is that near the original models, the interpolation actually leads to improved model performance. Concretely, given model A and permutationally aligned model B, the following simple recipe leads to an increase in test accuracy: mix weights linearly according to $(1 - \lambda) A + \lambda B$, for $\lambda \approx 0.1$ or symmetrically, $\lambda \approx 0.9$. We refer to this effect as BUMP, coined as a backronym for Brief Unexpected Mid-Interpolation Peak.
For various models, datasets, and network width multipliers, we demonstrate small, but statistically highly significant increases in test accuracy when applying such a recipe. While the gains are modest, they are highly consistent, warranting careful analysis and providing insights into merging and ensembling of deep neural networks, and motivating new model combination strategies.

Our contributions are as follows:\footnote{Code is available in the supplementary material, and will be released in a public repository upon acceptance.}
\vspace{-4pt}
\begin{itemize}
\item We empirically demonstrate that the BUMP phenomenon appears consistently across multiple datasets, architectures, training regimes, and persists under model degradation and at early checkpoints.
\item We formulate several plausible regularization-based hypotheses for the underlying causes of the BUMP phenomenon and empirically falsify them.
\item We analyze the BUMP from multiple perspectives, including its connection to classical logit ensembling and an inspection of model predictions.
\item We propose a novel ensembling method that combines weight mixing and logit mixing, achieving higher accuracy than both logit ensembling and the BUMP itself.
\end{itemize}

\section{Related Work}

\subsection{Linear mode connectivity and model re-basing}

Linear mode connectivity (LMC) is a widely studied phenomenon that provides insight into neural network loss landscapes and generalization. An intriguing direction considers permutation symmetries, \citet{entezari2022permutation} conjectured that, once these symmetries are accounted for, SGD solutions from different initializations become linearly mode connected. This conjecture was empirically demonstrated in \citet{ainsworth2022rebasin}, using a simple greedy neuron-matching algorithm family dubbed Git Re-Basin, but only for networks employing Layer Normalization \citep{ba2016layer} and very wide architectures (16x–32x wider than the baseline). \citet{jordan2022repair} later extended these results to networks using Batch Normalization \citep{ioffe2015batch}, and to networks without normalization. \citet{theus2025generalized} considers a broader class of symmetries (including permutations as a special case) and provides the first demonstration of LMC for transformers.

\subsection{Model merging and ensembling} 

Ensembling and model merging both aim to combine multiple trained models to improve performance, robustness, or generalization, but they differ in the level at which models are combined.

In ensembling, multiple base predictors are combined into a single predictor, ranging from simple output aggregation (e.g., averaging logits or probabilities) to more general learned combinations. Some methods aggregate model probabilities (soft voting), which work well for well-calibrated and confident models \cite{lakshminarayanan2017simple}. Others average logits, preserving relative confidence and often producing sharper, more accurate predictions \cite{Tassi2022, buchanan2023effects}. 

In model merging, models are combined by operating directly on network weights. Many approaches have been explored, \citet{izmailov2018averaging} introduces Stochastic Weight Averaging (SWA), which improves generalization by averaging points along an SGD trajectory using cyclical or constant learning rates. \citet{wortsman2022model} proposes model soups, where several fine-tuned models (that share a common initialization but are further trained with different hyperparameters) are weight-averaged to boost accuracy and robustness, offering a cheaper inference-time alternative to logit ensembling. Other methods use weighted combinations of models in federated learning \citep{mcmahan2017communication}. Another line of work, such as Fisher merging, uses Fisher information to guide the merging procedure \citep{matena2022merging}.

\section{Notations and formal definition of BUMPs}

Let $A$ and $B$ be two models with identical architectures and $d \in \mathbb{N}$ parameters, trained from different random initializations and batch orders. Let $\Theta_A \in \mathbb{R}^d$ and $\Theta_B \in \mathbb{R}^d$ denote their converged weight vectors. With a slight abuse of notation, we identify models with their weight vectors: we write $\Theta_A$ (resp.\ $\Theta_B$) to denote model $A$ (resp.\ $B$) with parameters set to $\Theta_A$ (resp.\ $\Theta_B$).

For a model $M$, datapoint $x\in \mathcal{D}$ and class $c\in \mathcal{C}$, we denote the logit vector corresponding to $x$ by $\ell_M(x)$, and denote the logit value corresponding to $x$ and class $c$ by $\ell_M(x, c)$, where $\mathcal{D}$ denotes the (input) dataset and $\mathcal{C}$ the set of class labels. Given models $\Theta_A, \Theta_B$ and coefficient $\lambda \in [0, 1]$, we define the \textbf{LogitMix} model as the classical logit ensemble of $\Theta_A$ and $\Theta_B$ with coefficient $\lambda$. We denote the resulting logit corresponding to $x$ by $\ell_{\text{LogitMix}}(A, B, x, \lambda) \coloneq \lambda\cdot \ell_B(x) + (1 -\lambda)\cdot \ell_A(x)$. We define the \textbf{WeightMix} as the weight-interpolated model $M(\lambda, \Theta_A, \Theta_B)$ with parameter vector $\lambda \Theta_B + (1 - \lambda) \Theta_A$, and denote its logits by $\ell_{\text{WeightMix}}(A, B, x, \lambda) \coloneq \ell_{M(\lambda, \Theta_A, \Theta_B)}(x)$.

\subsection{Loss barrier: worst performing model along the interpolation line}

Linear mode connectivity and the approximate convexity of the loss basin is captured through the notion of loss barrier.

\begin{definition}[Loss barrier \citep{frankle2020linear}]
Given two models $\Theta_A$ and $\Theta_B$ with $\mathcal{L}(\Theta_A) \approx \mathcal{L}(\Theta_B)$, the loss barrier is $\max_{\lambda \in [0,1]} \mathcal{L}((1-\lambda) \Theta_A + \lambda \Theta_B) - \frac{1}{2}(\mathcal{L}(\Theta_A) + \mathcal{L}(\Theta_B))$, where $\mathcal{L}$ is the utilized loss function.
\end{definition}

Two networks are considered linear mode connected if this barrier is small (near zero). Analogously, an accuracy barrier can be defined by comparing the minimum accuracy along the interpolation line to the average accuracy of $\Theta_A$ and $\Theta_B$. 

\citet{ainsworth2022rebasin} demonstrates zero-barrier linear mode connectivity between $\Theta_A$ and $\pi(\Theta_B)$, where $\pi(\Theta_B)$ is obtained by applying a neuron permutation to $\Theta_B$ that is computed to align it with $\Theta_A$. We call such $\Theta_A, \pi(\Theta_B)$ pairs ``aligned'' and refer to unpermuted $\Theta_A, \Theta_B$ pairs as ``naïve''.

\subsection{BUMP models: best performing model along the interpolation line}

In this work, we focus on the \textbf{best-performing} interpolated model, instead of the worst-performing one. Empirically, we consistently observe models near $A$ and $B$ whose accuracy exceeds that of the closer endpoint. Based on the shape of these interpolation accuracy curves (see Figure~\ref{fig:bumpwm16}), we call these \textbf{BUMP} models.

\begin{definition}[BUMP model]
Let $\Theta_A,\Theta_B \in \mathbb{R}^d$ be two models and let $\lambda\in[0,1]$. 
We call the WeightMix model $M(\lambda, \Theta_A, \Theta_B)$ a \emph{BUMP model} if its accuracy exceeds that of the closer endpoint, i.e.,
\[
\mathrm{acc}(M(\lambda, \Theta_A, \Theta_B)) >
\begin{cases}
\mathrm{acc}(\Theta_A), & \text{if } \lambda < \tfrac{1}{2},\\
\mathrm{acc}(\Theta_B), & \text{if } \lambda \ge \tfrac{1}{2}.
\end{cases}
\]
\end{definition}

\section{Empirical demonstration of the BUMP}\label{sec:bumpdemo}

In this section, we provide empirical evidence that the BUMP is a consistently occurring phenomenon. We experiment on multiple datasets and architectures. We report results on CIFAR-10~\citet{krizhevsky2009learning}, training $100$ ResNet-20~\citet{resnet} models with different initialization seeds, for all width multiplier in $\{4, 8, 16\}$. We form $50$ model pairs, and align each pair via the weight-matching algorithm of \citet{ainsworth2022rebasin}. We interpolate linearly between the aligned models using 25 equally spaced coefficients in $[0, 1]$. We also train Tiny-10 models on CIFAR-10, and ResNet-20 models on CIFAR-100~\citet{krizhevsky2009learning}, the results can be found in Appendix~\ref{appendix:tiny10} and~\ref{appendix:cifar100}. The training details for both model architectures can be found in Appendix~\ref{appendix:training_details}. (For additional evidence with further models and training regimes, see Section~\ref{sec:robust}.)

\begin{figure}
\centering
    \subfloat[Left BUMP (aligned)]{%
            \includegraphics[trim=0cm 0cm 0cm 0cm, clip=true, width=0.5\columnwidth]{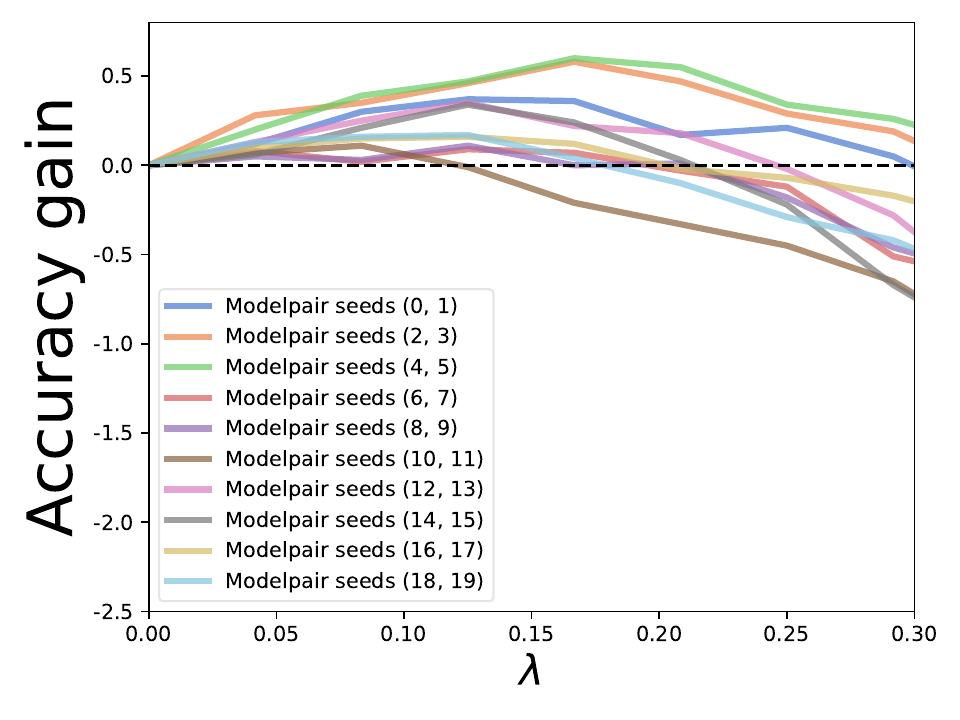}%
        \label{fig:bumpleftwm16}%
        }%
    \subfloat[Right BUMP (aligned)]{%
        \includegraphics[trim=0cm 0cm 0cm 0cm, clip=true, width=0.5\columnwidth]{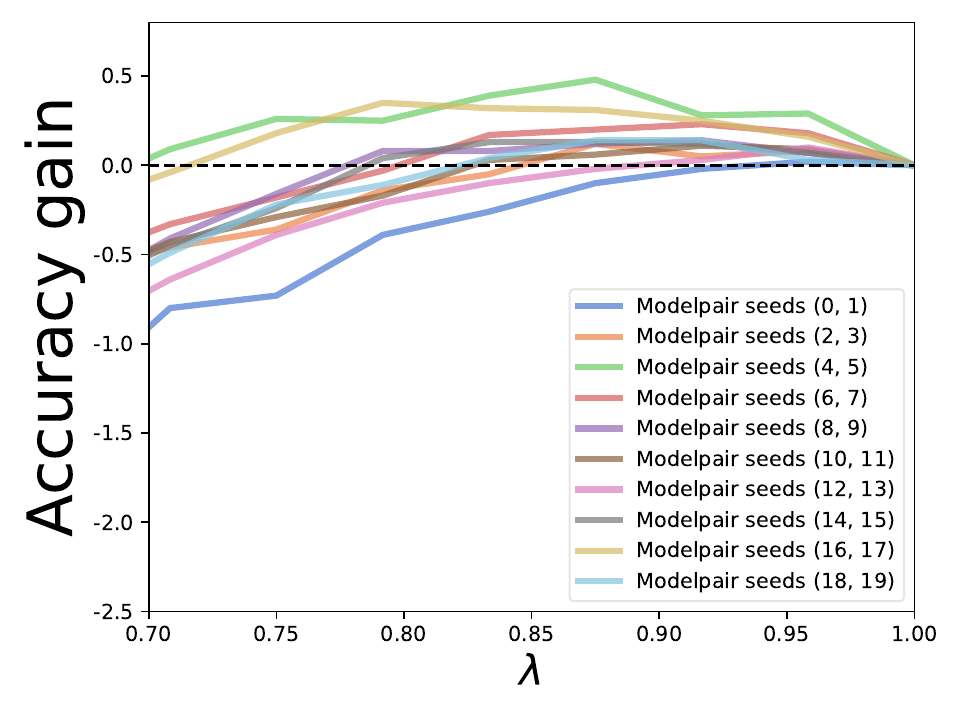}%
        \label{fig:bumprightwm16}%
        }%
    \caption{The two BUMPs when interpolating between aligned models, the accuracies are measured on the CIFAR-10 test set, models are ResNet-20, showing $10$ model pairs. Width multiplier is 16. We shifted the plots so that all accuracies of the original models are 0.}%
    \label{fig:bumpwm16}%
\end{figure}

\begin{figure}[t]
\centering
    \subfloat[Left BUMP (na\"ive)]{%
        \includegraphics[trim=0cm 0cm 0cm 0cm, clip=true, width=0.5\columnwidth]{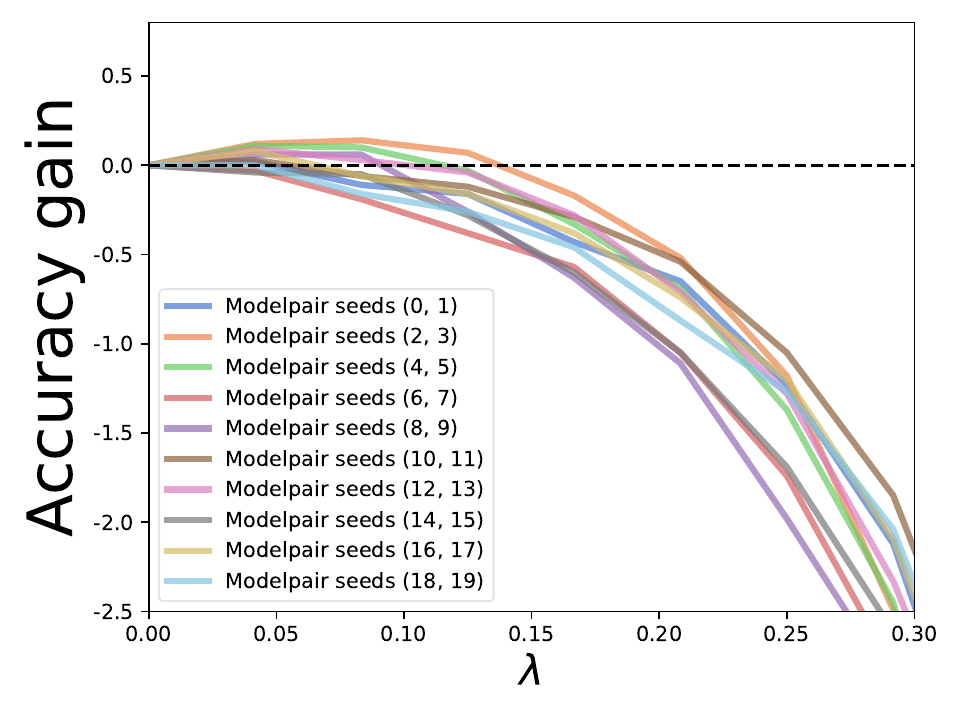}%
        \label{fig:bumpnaiveleftwm32}%
        }%
    \subfloat[Right BUMP (na\"ive)]{%
        \includegraphics[trim=0cm 0cm 0cm 0cm, clip=true, width=0.5\columnwidth]{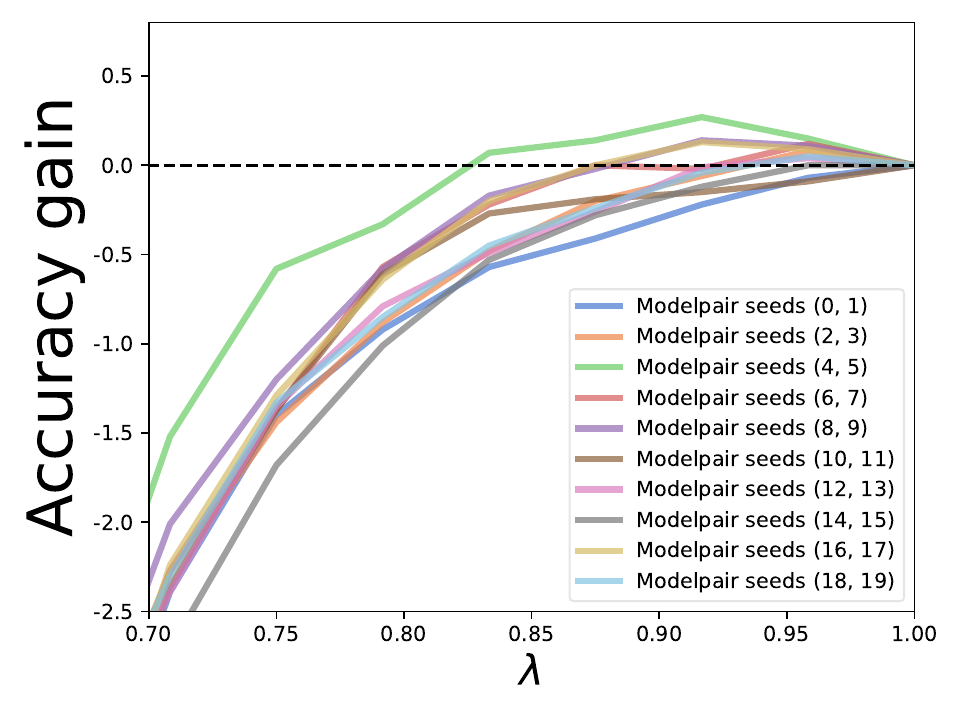}%
        \label{fig:bumpnaiverightwm32}%
        }%
    \caption{The two BUMPs when interpolating between models without alignment, the accuracies are measured on the CIFAR-10 test set, models are ResNet-20, showing $10$ model pairs. Width multiplier is 16. We shifted the plots so that all accuracies of the original models are 0.}%
    \label{fig:bumpnaivewm16}%
\end{figure}

For $16$x-width, we plot interpolation curves for $\lambda \in (0.0, 0.3)$ and $\lambda \in(0.7, 1.0)$ across $10$ model pairs. Figure~\ref{fig:bumpwm16} shows clear BUMPs near both endpoints for aligned model pairs, whereas for naïve pairs the BUMP is absent or much smaller (Figure~\ref{fig:bumpnaivewm16}). For width multipliers $1$ and $2$, BUMPs are negligible (note, that for these widths, the loss barriers remain large even after permutation). BUMPs also appear at $32$x-width, but with smaller accuracy gains than at $16$x-width. 

To quantify the accuracy gain of a BUMP we use multiple methods. We denote a model $M$'s training and test accuracies by $\text{acc}_\text{train}(M)$ and $\text{acc}_\text{test}(M)$ respectively. 

\paragraph{Method ``Fixed-$\lambda$'':}

For a model pair $(\Theta_A, \Theta_B)$, we fix the interpolation coefficient $\lambda$ to obtain models $M_1 = M(\lambda, \Theta_A, \Theta_B)$ and $M_2 = M(1 - \lambda, \Theta_A, \Theta_B)$, i.e., we evaluate BUMPs at $\lambda$ and $1-\lambda$. The $\lambda$ parameter was chosen based on extensive experiments on the \textit{training dataset} (determining highest accuracy BUMP locations and averaging the results). Based on these evaluations we chose $\lambda = 0.05$ for $4$x-width, $\lambda = 0.0075$ for $8$x-width and $\lambda = 0.1$ for $16$x-width. We measure BUMP accuracy gains as $\text{acc}_\text{test}(M_1) - \text{acc}_\text{test}(A)$ and $\text{acc}_\text{test}(M_2) - \text{acc}_\text{test}(B)$.

\paragraph{Method ``Test'':} For a model pair ($\Theta_A, \Theta_B$) we search for the interpolated model with the highest accuracy on the test dataset separately on the intervals $[0, 0.5)$ and $[0.5, 1]$, yielding two models $M_1$ and $M_2$. Again, we measure BUMP accuracy gains as $\text{acc}_\text{test}(M_1) - \text{acc}_\text{test}(A)$ and $\text{acc}_\text{test}(M_2) - \text{acc}_\text{test}(B)$. We are aware that methodologically this constitutes data contamination (as we are selecting the highest accuracy on the test set), but we include these values because the smoothness of the accuracy curves suggests that they are more indicative of the accuracy gains achievable by the method.

Since the models $A$ and $B$ play symmetric roles, we average all $100$ BUMP gains per \emph{dataset--architecture--width} setting, rather than treating the first and second BUMPs separately ($50$ each). We report results for both methods on CIFAR-10, ResNet-20 in Table~\ref{tab:bumppermc10r20} (aligned pairs) and Table~\ref{tab:bumpnaivec10r20} (naïve pairs). These demonstrate that BUMP gains are consistently present, and that alignment substantially amplifies the BUMP effect compared to naïve pairs.

\begin{table}[h!]
\centering
\caption{
    Average accuracy gains of the BUMPs over their corresponding endpoint models on \textbf{aligned model pairs}. Results are reported on the CIFAR-10 test set using aligned ResNet-20 pairs (100 models) trained to convergence (250 epochs). Standard deviations are denoted in parentheses.
}
\begin{tabular}{lcc}
\toprule
\multicolumn{3}{c}{Accuracy gain \% --- Aligned pairs} \\
\cmidrule(lr){1-3}
Width & Fixed-$\lambda$ & Test \\
\midrule
4x-width  & $0.053\,{\scriptstyle(\pm 0.096)}$ & $0.087\,{\scriptstyle(\pm 0.081)}$ \\
8x-width  & $0.118\,{\scriptstyle(\pm 0.111)}$ & $0.164\,{\scriptstyle(\pm 0.117)}$ \\
16x-width & $0.181\,{\scriptstyle(\pm 0.137)}$ & $0.227\,{\scriptstyle(\pm 0.140)}$ \\
\bottomrule
\end{tabular}

\label{tab:bumppermc10r20}
\end{table}

\begin{table}[ht!]
\centering
\caption{
    Average accuracy gains of the BUMPs over their corresponding endpoint models on \emph{non-aligned}, i.e., \textbf{naïve model pairs}. Results are reported on the CIFAR-10 test set using aligned ResNet-20 pairs (100 models) trained to convergence (250 epochs). Standard deviations are denoted in parentheses.
}
\begin{tabular}{lcc}
\toprule
\multicolumn{3}{c}{Accuracy gain \% --- Naïve pairs} \\
\cmidrule(lr){1-3}
Width & Fixed-$\lambda$ & Test \\
\midrule
4x-width  & $-0.041\,{\scriptstyle(\pm 0.111)}$ & $0.038\,{\scriptstyle(\pm 0.061)}$ \\
8x-width  & $-0.041\,{\scriptstyle(\pm 0.123)}$ & $0.051\,{\scriptstyle(\pm 0.065)}$ \\
16x-width & $-0.073\,{\scriptstyle(\pm 0.147)}$ & $0.056\,{\scriptstyle(\pm 0.069)}$ \\
\bottomrule
\end{tabular}

\label{tab:bumpnaivec10r20}
\end{table}

\section{Robustness of the BUMP phenomenon}\label{sec:robust}

In this section, we show that the BUMP phenomenon is robust and widespread: it can be observed consistently across architectures and training regimes, and it persists under substantial degradation of the underlying models. Concretely, we provide evidence along three axes: (i) the effect is visible (though typically unremarked) in interpolation figures of prior works, (ii) it already appears well before full convergence along the training trajectory and is often \emph{more pronounced at earlier checkpoints} than at convergence, and (iii) it survives—and frequently \emph{strengthens}—under extreme sparsification via magnitude pruning. Together, these results argue against the interpretation of BUMPs as fragile interpolation artifacts that occur only under carefully tuned conditions.

\subsection{BUMPs in prior works}

After observing the BUMP phenomenon in our own experiments, we examined prior works, and found similar local performance improvements near the endpoints of aligned interpolation curves across a wide range of studies, including those involving convolutional networks and transformers. While these effects are visible in several published interpolation plots, they are not discussed and are at most briefly mentioned; in other words, the BUMP phenomenon is not explicitly explored in these works (see Figure 1 in \citet{mcmahan2017communication}, Figure 5 in \citet{ainsworth2022rebasin} and Figure 4 in \citet{theus2025generalized}).

\subsection{BUMPs at earlier checkpoints}

To test whether BUMPs depend on full convergence, we analyze interpolations between models taken from earlier checkpoints of the training process.

\begin{figure}[tbp]
  \centering
  \includegraphics[width=0.98\columnwidth]{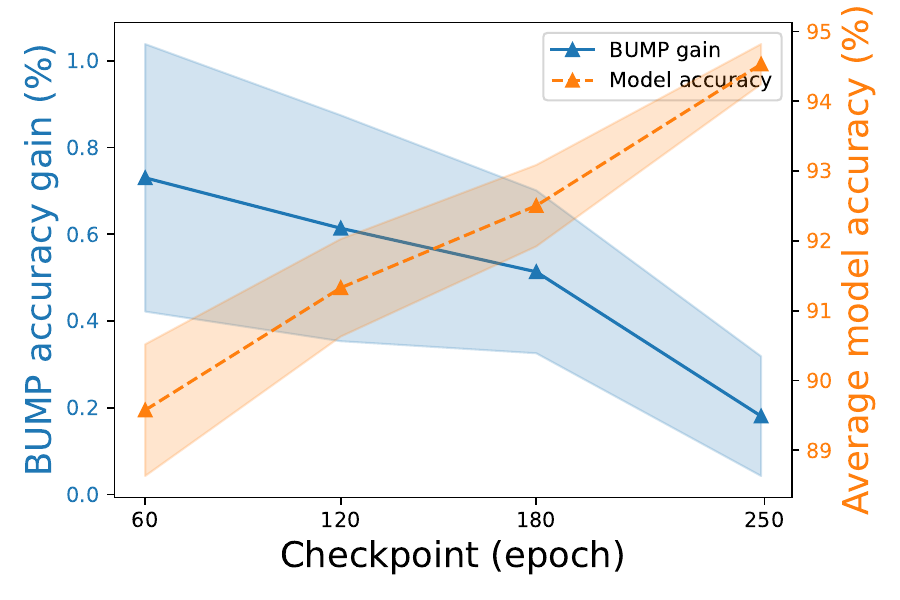}  
  \caption{BUMP gain (solid line, left axis) and average model accuracy (dashed line, right axis) as a function of training epochs for $16$x-width ResNet-20 models on CIFAR-10. Values are averages for $50$ model pairs, bands represent standard deviations.}
  \label{fig:checkpoint_bumps}
\end{figure}

Figure~\ref{fig:checkpoint_bumps} show the results for checkpoints 60, 120, 180, and the full training of 250 epochs. We can observe that BUMPs are already present well before convergence and their magnitude is often larger at intermediate checkpoints than for fully trained models, and diminishes toward convergence. This shows that the phenomenon does not rely on late-stage optimization and can be strongest when models are still relatively inaccurate.

\subsection{BUMPs in pruned models}

We study how weight pruning affects the BUMP phenomenon, asking whether BUMPs persist under extreme sparsification. We apply a single round of global magnitude pruning whose purpose is not pruning optimization but deliberate performance degradation and sparsification. For the $16$x-width ResNet-20 model on CIFAR-10, we prune $95\%$ of the weights and evaluate interpolations using the same Fixed-$\lambda$ protocol as in Section~\ref{sec:bumpdemo}.

We consider two approaches: (i) pruning first, followed by alignment via weight permutation, and (ii) alignment first, followed by pruning. In both cases, BUMPs remain clearly present. When pruning is applied before alignment, the average Fixed-$\lambda$ BUMP gain is $1.54\% \pm 1.07\%$; when pruning is applied after alignment, the gain is $1.60\% \pm 1.11\%$. Notably, these gains are substantially larger than those observed for fully dense, converged models, despite pruning reducing average test accuracy from approximately $94.5\%$ to $74.6\%$. This demonstrates that BUMPs are robust to severe sparsification and do not rely on dense or carefully optimized model representations.

Overall, these results establish the BUMP phenomenon as a systematic and robust property of neural network interpolations, persisting across architectures, training stages, and even under severe model degradation.

\section{The BUMP is a result of a simple regularization effect?}
\label{sec:regularization}

When seeking for an explanation for the BUMP phenomenon, a natural hypothesis is that BUMP reflects a simple form of regularization. We investigate this hypothesis from multiple perspectives, including whether the effect depends on moving toward another trained model or whether it can be reproduced by adding noise to the logits.

\subsection{Is the direction important?}

In the previous experiments, we moved in parameter space from one trained model toward another. This raises the question of whether the BUMP specifically depends on moving toward another trained model. To test this, we conduct two sets of experiments.

\subsubsection{Random direction model}

In this experiment, we interpolate between a trained model $A$ and a random model $R$ with parameter vector $\Theta_R$ sampled uniformly from a hypersphere centered at $A$. For each aligned pair $(A, \pi(B))$ we sample $R$ on the sphere with radius $\|\Theta_A - \Theta_B\|_2$. To measure accuracy change, we use slightly modified versions of our earlier methods. With Fixed-$\lambda$, we evaluate only the model $(1 - \lambda) \Theta_A + \lambda \Theta_R$, and for method ``Test'', we search over the full interval $[0, 1]$ to obtain the interpolating coefficient, instead of splitting into halves. Table~\ref{tab:random_r20c10} reports results for width multipliers $4, 8, 16$, averaged over model pairs. 

\begin{table}[h!]
\centering
\caption{
    Average accuracy differences when moving towards \textbf{random directions} on CIFAR-10 with ResNet-20 ($50$ model pairs), with standard deviations denoted in parentheses.
}
\begin{tabular}{lcc}
\toprule
\multicolumn{3}{c}{Accuracy gain \% --- Random direction} \\
\cmidrule(lr){1-3}
Width & Fixed-$\lambda$ & Test \\
\midrule
4x-width  & $-0.009\,{\scriptstyle(\pm 0.065)}$ & $0.028\,{\scriptstyle(\pm 0.035)}$ \\
8x-width  & $-0.012\,{\scriptstyle(\pm 0.056)}$ & $0.035\,{\scriptstyle(\pm 0.040)}$ \\
16x-width & $-0.002\,{\scriptstyle(\pm 0.053)}$ & $0.031\,{\scriptstyle(\pm 0.035)}$ \\
\bottomrule
\end{tabular}
\label{tab:random_r20c10}
\end{table}

We find no BUMPs when using a Fixed-$\lambda$, and the BUMPs selected by the ``Test'' method are negligible compared to those observed when interpolating toward $\pi(B)$. 

\begin{figure}
\centering
    \subfloat[Distance multiplier 1]{%
        \includegraphics[trim=0cm 0cm 0cm 0cm, clip=true, width=0.45\columnwidth]{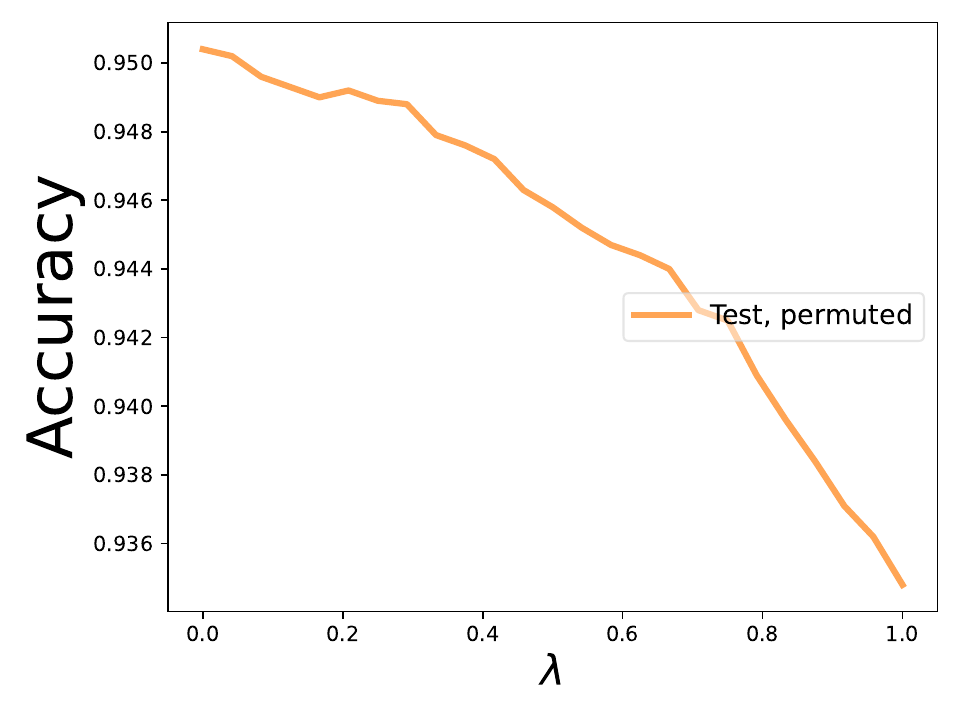}%
        \label{fig:random1}%
        }%
    \hfill%
    \subfloat[Distance multiplier 5]{%
        \includegraphics[trim=0cm 0cm 0cm 0cm, clip=true, width=0.45\columnwidth]{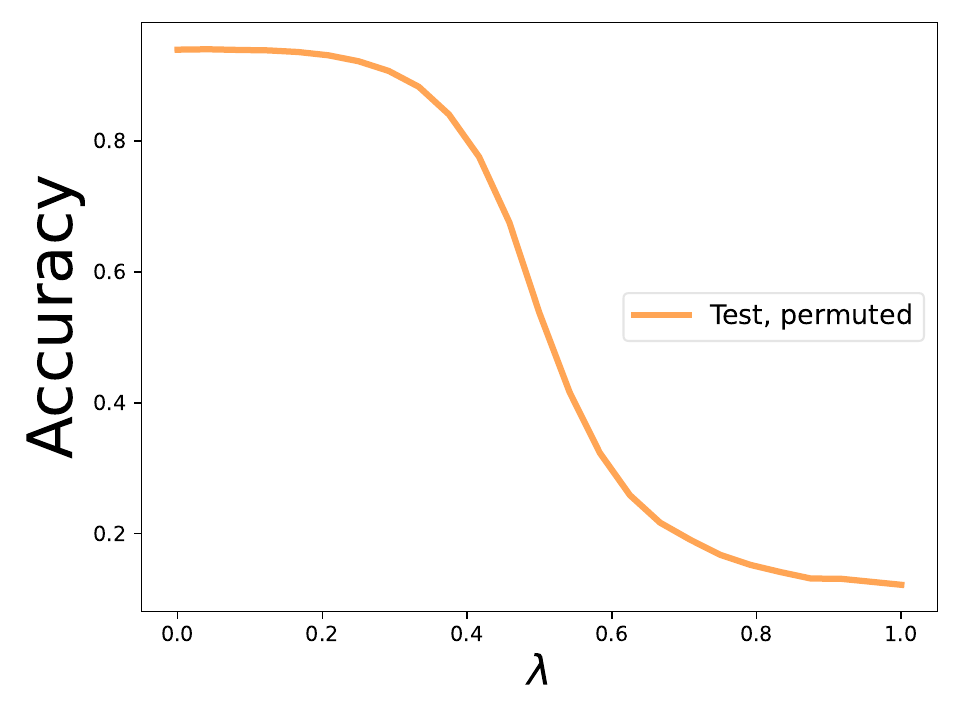}%
        \label{fig:random5}%
        }%
    \caption{Interpolation in the direction of a randomly chosen model on CIFAR-10, ResNet-20. Width multiplier is 16, the distance multiplier determines the distance of model $A$ from the random model $R$ compared to the distance of model $A$ and $B$.}%
    \label{fig:random}%
\end{figure}

Consistent with this, the interpolation line segment between $A$ and a randomly sampled model shows only minor accuracy variation (Figure~\ref{fig:random1}). When we increase the sphere radius to $5 \cdot \|\Theta_A - \Theta_B\|_2$, accuracy stays nearly constant for a short range and then drops sharply (Figure~\ref{fig:random5}).

\subsubsection{Randomly reflected model}

To further test whether moving toward a trained model is essential, we construct a perturbed version of $\pi(B)$ by randomly ``reflecting'' parameters. Specifically, for each parameter pair $(w_a, w_b) \in \Theta_A \times \pi(\Theta_B)$, we set $w_b \coloneq 2w_a - w_b$ with probability $0.5$, yielding a new model $B'$ that is, coordinate-wise, equally far from $A$. We then interpolate between $A$ and $B'$. Table~\ref{tab:randomopp_r20c10} reports results averaged over all model pairs. As in the previous experiment, we observe no significant accuracy gains. 

\begin{table}[h!]
\centering
\caption{
    Average accuracy differences when moving towards models with \textbf{randomly reflected} weights on CIFAR-10 with ResNet-20 ($50$ model pairs). Standard deviations are denoted in parentheses.
}
\begin{tabular}{lcc}
\toprule
\multicolumn{3}{c}{Accuracy gain \% --- Randomly reflected} \\
\cmidrule(lr){1-3}
Width & Fixed-$\lambda$ & Test \\
\midrule
4x-width  & $-0.081\,{\scriptstyle(\pm 0.117)}$ & $0.025\,{\scriptstyle(\pm 0.042)}$ \\
8x-width  & $-0.094\,{\scriptstyle(\pm 0.112)}$ & $0.022\,{\scriptstyle(\pm 0.043)}$ \\
16x-width & $-0.077\,{\scriptstyle(\pm 0.110)}$ & $0.028\,{\scriptstyle(\pm 0.046)}$ \\
\bottomrule
\end{tabular}
\label{tab:randomopp_r20c10}
\end{table}

\subsection{Adding noise to the logits}

\begin{figure}[tbp]
  \centering
  \includegraphics[width=0.98\columnwidth]{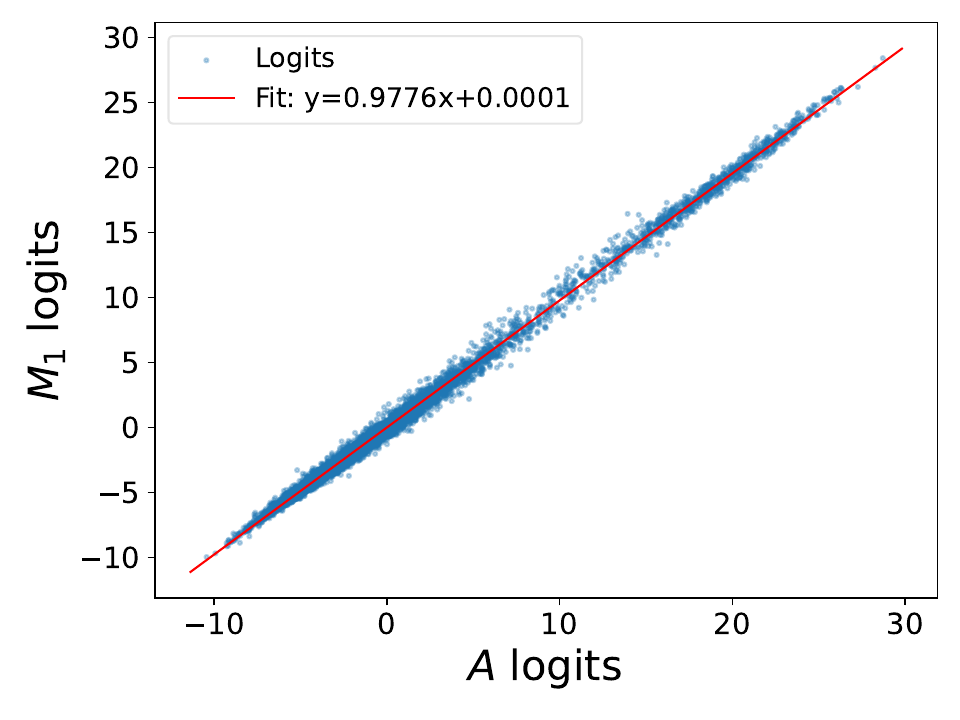}  
  \caption{Relationship between logit values of $A$ and $M_1$ models on CIFAR-10, ResNet-20, width multiplier is $16$. The red line indicates the best-fit affine relationship, highlighting the linear correspondence between the logits of the original and BUMP models.}
  \label{fig:logit_relationship}
\end{figure}

The hypothesis that the BUMP is a regularization effect is particularly compelling in light of the relationship between the logits produced by model $A$ and those produced by model $M_1$ (the closer BUMP to $A$). As shown in Figure \ref{fig:logit_relationship}, there is an extremely strong linear relationship across data points and labels: to a first approximation, each logit output by model $M_1$ is simply a constant multiple of the corresponding logit from model $A$. This might suggest a seemingly trivial explanation, namely that such a constant rescaling alone improves performance. However, this cannot be the case, since scaling logits by a constant factor does not change the predicted labels.

We now exclude two, successively more subtle variations of this hypothesis, namely that the logits of model $M_1$ can be explained as a noisy affine transformation of the logits of model $A$.

\subsubsection{Homoscedastic affine noise model}

In this experiment, we model the conditional distribution of $\ell_{M_1}(x, c)$ given $\ell_A(x, c)$ as $\ell_{M_1}(x, c) = a\cdot \ell_A(x, c) + b + \varepsilon$, with scalar parameters $a, b, \sigma \in \mathbb{R}$, where $\varepsilon\sim \mathcal{N}(0, \sigma^2)$. The parameters $a, b$ are fitted by least squares, and the $\sigma$ is the empirical standard deviation of $\ell_{M_1}(x, c) - (a\cdot \ell_A(x, c) + b)$ on the test dataset. We then replace the true logit values $\ell_{M_1}(x, c)$ by samples drawn from the fitted conditional distribution and evaluate classification accuracy using these synthetic logits. If model $M_1$ differed from model $A$ only by an affine transformation with homoscedastic noise, sampling from this model would be expected to preserve or improve accuracy. However, we observe no statistically significant improvement, see Table~\ref{tab:noise_model_gains} in Appendix~\ref{appendix:cifar10_moretables}.

\subsubsection{Heteroscedastic affine noise model}

We next consider a more flexible model in which the variance of the noise depends on the logit value of model $A$. Specifically, we set $\ell_{M_1}(x, c) = a\cdot \ell_A(x, c) + b + \varepsilon$, where $\varepsilon\sim \mathcal{N}(0, \sigma(\ell_A(x, c))^2)$, allowing the noise to vary as a function of $\ell_A(x, c)$. The parameters $a, b$ are again fitted by least squares, while $\sigma$ is estimated by binning the values of $\ell_A(x, c)$, estimating the empirical variance within each bin and fitting a polygonal line on these values. We used $10$ bins in our experiments. Again, we replace the true logit values with samples and we observe significantly worse results in terms of accuracy, see Table~\ref{tab:noise_model_gains}  in Appendix~\ref{appendix:cifar10_moretables}.

\section{Comparing BUMP to Classical Logit Ensembling}\label{sec:ensanalysis}

\begin{figure}[tbp]
  \subfloat[Accuracy curves]{%
    \centering
    \includegraphics[width=0.48\columnwidth]{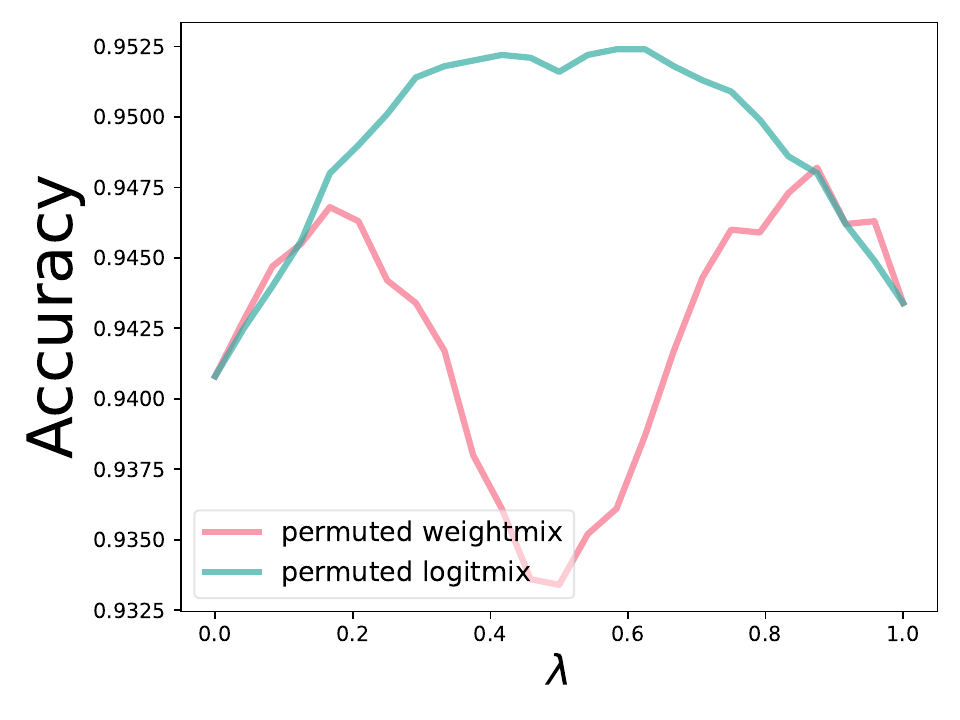}
      \label{fig:logitweight}
  }%
  \subfloat[KL divergences]{
    \includegraphics[width=0.48\columnwidth]{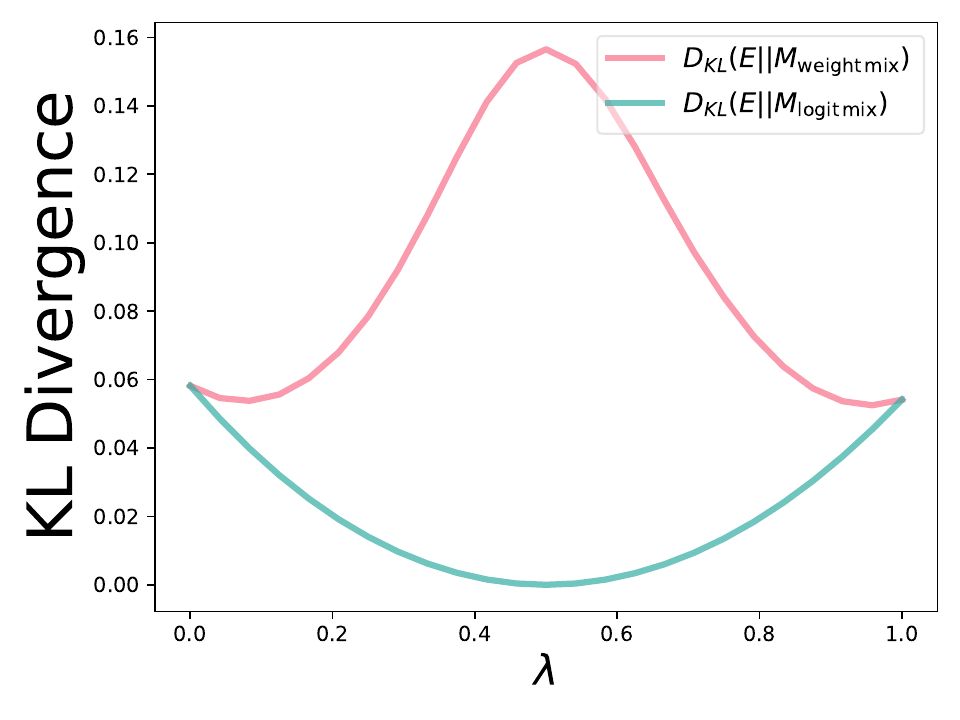}
      \label{fig:kl_div_wmlmE}
  }%
  \caption{\textbf{Left (comparing accuracy curves):} The WeightMix and LogitMix accuracy curve on CIFAR-10, ResNet-20, width multiplier is $16$. \textbf{Right (comparing class probabilities):} Kullback-Leibler divergence between the class probability distributions of the WeightMix/LogitMix models and the $\lambda=0.5$ LogitMix (Ensemble) model, CIFAR-10, ResNet-20, width multiplier is $16$.}
\end{figure}

In this section, we investigate how the BUMP phenomenon relates to classical logit ensembling, i.e., LogitMix. Figure~\ref{fig:logitweight} compares the accuracy curves of the LogitMix and WeightMix models for two $16$x-width models.  Although the two curves behave similarly near the endpoints, this similarity alone does not imply that the underlying mechanisms are the same. To address this question, we carry out a geometric and probabilistic analysis comparing the two models in detail.

\subsection{Inspecting the angles}

We directly compare the logits of the WeightMix and LogitMix models. More precisely, we measure the angle between the logits for the same inputs. If the two mixing procedures created identical logits, this angle would be zero.

\begin{figure}[tbp]
  \centering
  \includegraphics[width=0.98\columnwidth]{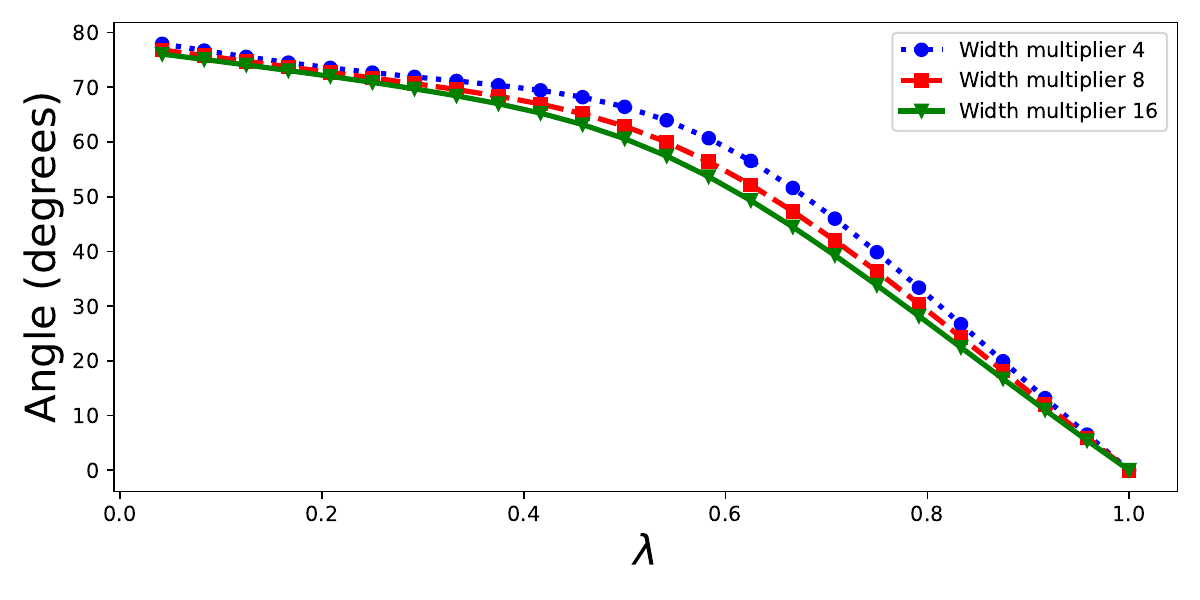}  
  \caption{The mean angles across the test datapoints for $g = 24$, CIFAR-10, ResNet-20, width multiplier is 16. Values are averages for $50$ model pairs.}
  \label{fig:meanangles}
\end{figure}

For a model pair $(A, B)$, $g\in \mathbb{Z}^{+}$ and a datapoint $x$, let 
\[
v_{\ell}(x) \coloneqq \ell_{\text{LogitMix}}(A, B, x, 1) - \ell_{\text{LogitMix}}(A, B, x, 0),
\]
\[
v_{w, i}(x) \coloneqq \ell_{\text{WeightMix}}(A, B, x, i/g) - \ell_{\text{WeightMix}}(A, B, x, 0),
\]
for $i = 1, \dots, g$. (Note, that since we are comparing the angles, and LogitMix logits all lie on the line segment determined by the logits corresponding to models $A, B$, we may choose one of the LogitMix logit differences as a reference.) For each $i$, we compute the angle between $v_{\ell}(x)$ and $v_{w, i}(x)$, average over test datapoints, and then average over the $50$ model pairs. Figure~\ref{fig:meanangles} reports the mean angle as a function of $\lambda = i/g$. (For $i = g$, we recover model $B$, so the angle is necessarily $0^\circ$.) The large angles for small $i$ indicate that WeightMix and LogitMix models produce substantially different logits near model $A$. 

\subsection{Probability-Space Proximity to the Logit Ensemble}

To compare weight mixing and logit ensembling at the granularity of predicted class-distributions, we take the $\lambda=0.5$ LogitMix model as a reference, denoted by $E$ (for \emph{Ensemble}), and measure with the Kullback--Leibler divergence how closely WeightMix and LogitMix track it. Figure~\ref{fig:kl_div_wmlmE} shows the results. Near the endpoints $A$ and $B$, the divergence decreases (even for WeightMix models), indicating that the interpolation path moves toward $E$ rather than away from it. However, Figure~\ref{fig:kl_div_wmlmE} also makes clear that WeightMix approaches $E$ only modestly compared to LogitMix: the predictive distributions remain substantially farther from $E$ across $\lambda$. One possible interpretation is that both procedures move toward $E$ simply because the ensemble distribution concentrates better predictions, and many reasonable perturbations from $A$ or $B$ may initially improve agreement with it. The quantitative gap suggests that weight mixing is not equivalent to classical logit ensembling, instead, it induces a distinct transformation of the predictive distribution.

A natural question arises: is weight mixing still merely an approximation of classical ensembling, or does it provide complementary information that can be exploited beyond standard logit mixing?

\section{A novel ensembling method: combining weight and logit mixing}\label{sec:novelensemble}

In this section, we introduce \textbf{Weight–Prediction Mixed Ensembling (WPME)}, a novel ensembling method that jointly combines models in weight space and prediction space, and surpasses the accuracy of both classical logit ensembling and the BUMP itself. Classical ensembling combines model predictions (logits or probabilities) to reduce variance, but it does not leverage information directly contained in the model parameters. Our novel ensembling scheme demonstrates that weight mixing provides complementary information to output-space aggregation and can be exploited to achieve higher generalization performance.

\begin{definition}[Weight--Prediction Mixed Ensembling (WPME)]
Given two models $\Theta_A$, $\Theta_B$ and coefficients $\lambda,\beta \in \mathbb{R}$, we form the WeightMix model $M(\lambda,\Theta_A,\Theta_B)$.
For an input $x$, let $\ell_B(x)$ denote the logits of model $\Theta_B$, and let $\ell_{M(\lambda)}(x)$ denote the logits of $M(\lambda,\Theta_A,\Theta_B)$.
We define two variants:
\[
\begin{aligned}
\ell_{\mathrm{WPME\text{-}Logit}}(x;\lambda,\beta)
&= (1-\beta)\,\ell_{M(\lambda)}(x) + \beta\,\ell_B(x),\\[2pt]
p_{\mathrm{WPME\text{-}Prob}}(x;\lambda,\beta)
&= (1-\beta)\,\mathrm{softmax}\!\big(\ell_{M(\lambda)}(x)\big)\\
&\quad + \beta\,\mathrm{softmax}\!\big(\ell_B(x)\big).
\end{aligned}
\]
The prediction of WPME--Logit is obtained by applying the softmax to $\ell_{\mathrm{WPME\text{-}Logit}}$.
\end{definition}

Setting $\beta=0$ yields predictions from the WeightMix model $M(\lambda, \Theta_A, \Theta_B)$, and $\beta=1$ yields predictions from the endpoint model $B$. Thus $\beta$ controls the trade-off between the WeightMix model and the original predictor at endpoint B, rather than directly averaging the logits or weights of model $A$ and $B$.

\begin{figure}[t]
\centering
\begin{subfigure}{0.48\linewidth}
  \centering
  \includegraphics[width=\linewidth]{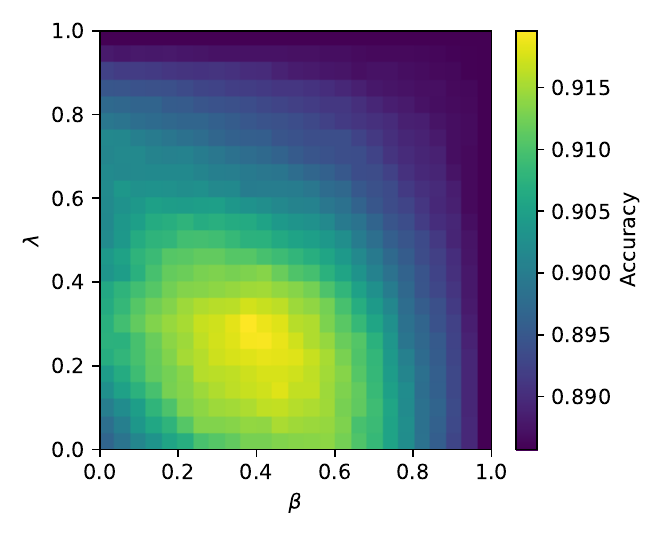}
  \caption{Checkpoint at epoch 60}
\end{subfigure}
\hfill
\begin{subfigure}{0.48\linewidth}
  \centering
  \includegraphics[width=\linewidth]{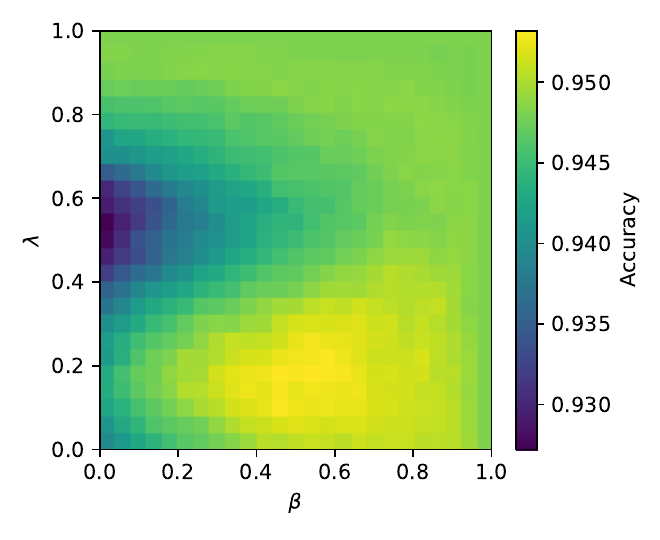}
  \caption{Checkpoint at convergence}
\end{subfigure}
\caption{
Accuracy over $(\lambda,\beta)$ for WPME obtained by interpolating model parameters in weight-space ($\lambda$)
followed by logit-space mixing ($\beta$) with endpoint model B. Results are shown for a partially trained model (left) and the fully trained model (right), CIFAR-10, ResNet-20, width multiplier is $16$. 
}
\label{fig:ensemble_acc_surface}
\end{figure}

\paragraph{WPME accuracy outperforms baselines.}
Figure~\ref{fig:ensemble_acc_surface} shows accuracy heatmaps obtained by sweeping the weight mixing coefficient $\lambda$
and the prediction mixing coefficient $\beta$.
Across both checkpoints, WPME exhibits broad regions of high accuracy that exceed those achieved by classical
logit mixing or the weight mixing (i.e., the BUMP). (Note, that classical logit mixing corresponds to the horizontal line $\lambda = 0$, while weight mixing to the vertical line $\beta=0$ on the heatmap.)
The accuracy landscape is smooth and unimodal, indicating that weight-space interpolation and
prediction-space interpolation provide complementary benefits rather than requiring precise hyperparameter tuning.

\paragraph{WPME improves the accuracy--calibration trade-off.}
As shown in Table~\ref{tab:ensemble}, WPME-logit achieves the highest accuracy at both checkpoints while modestly improving calibration relative to LogitMix; WPME-prob attains near-SoftMaxMix calibration with slightly higher accuracy (where SoftmaxMix simply linearly averages predicted probabilities, $p(x)=(1-\beta)\,\mathrm{softmax}(\ell_A(x))+\beta\,\mathrm{softmax}(\ell_B(x))$).

\begin{table*}[t]
\centering
\caption{
Comparison of Weight--Prediction Mixed Ensembling with baseline methods on CIFAR-10, ResNet-20, width-multiplier 16:
accuracy (Acc.), negative log-likelihood (NLL), and expected calibration error (ECE)
values for different ensembling methods of two models at two different stages of model
training (epoch 60 and 250). Reported values are sample means ± standard deviations (SD). 
Best results within each checkpoint are shown in bold.
}
\tiny
\setlength{\tabcolsep}{4pt}

\begin{minipage}[t]{0.48\textwidth}
\centering
\begin{tabular}{lccc}
\toprule
\multicolumn{4}{c}{\textbf{Model checkpoints: @60 epochs}}\\
\midrule
Method & Acc. $\uparrow$ & NLL $\downarrow$ & ECE $\downarrow$ \\
\midrule
A (orig) & 0.89972\,{$\scriptstyle(\pm 0.00871)$} & 0.30911\,{$\scriptstyle(\pm 0.03136)$} & 0.02789\,{$\scriptstyle(\pm 0.00650)$}\\
B (orig) & 0.90011\,{$\scriptstyle(\pm 0.01077)$} & 0.31030\,{$\scriptstyle(\pm 0.04081)$} & 0.02893\,{$\scriptstyle(\pm 0.00905)$} \\
LogitMix & 0.92369\,{$\scriptstyle(\pm 0.00591)$} & 0.22944\,{$\scriptstyle(\pm 0.01954)$} & 0.01374\,{$\scriptstyle(\pm 0.00385)$} \\
SoftMaxMix & 0.92160\,{$\scriptstyle(\pm 0.00607)$} & 0.24270\,{$\scriptstyle(\pm 0.01824)$} & 0.01990\,{$\scriptstyle(\pm 0.00482)$} \\
WeightMix & 0.91268\,{$\scriptstyle(\pm 0.00593)$} & 0.25942\,{$\scriptstyle(\pm 0.01875)$} & 0.01387\,{$\scriptstyle(\pm 0.00514)$} \\
\rowcolor{oursrow}
WPME-logit & \textbf{0.92605\,{$\scriptstyle(\pm 0.00513)$}} & \textbf{0.21918\,{$\scriptstyle(\pm 0.01612)$}} & \textbf{0.01015\,{$\scriptstyle(\pm 0.00297)$}} \\
\rowcolor{oursrow}
WPME-prob & 0.92424\,{$\scriptstyle(\pm 0.00567)$} & 0.23245\,{$\scriptstyle(\pm 0.01576)$} & 0.02124\,{$\scriptstyle(\pm 0.00682)$} \\
\bottomrule
\end{tabular}
\end{minipage}
\hfill
\begin{minipage}[t]{0.48\textwidth}
\centering
\begin{tabular}{lccc}
\toprule
\multicolumn{4}{c}{\textbf{Model checkpoints: @250 (full training)}}\\
\midrule
Method & Acc. $\uparrow$ & NLL $\downarrow$ & ECE $\downarrow$ \\
\midrule
A (orig) & 0.94949\,{$\scriptstyle(\pm 0.00349)$} & 0.23465\,{$\scriptstyle(\pm 0.01756)$} & 0.03313\,{$\scriptstyle(\pm 0.00256)$} \\
B (orig) & 0.94968\,{$\scriptstyle(\pm 0.00308)$} & 0.23454\,{$\scriptstyle(\pm 0.01448)$} & 0.03315\,{$\scriptstyle(\pm 0.00243)$} \\
LogitMix & 0.95790\,{$\scriptstyle(\pm 0.00262)$} & 0.18520\,{$\scriptstyle(\pm 0.01103)$} & 0.02598\,{$\scriptstyle(\pm 0.00222)$} \\
SoftMaxMix & 0.95742\,{$\scriptstyle(\pm 0.00288)$} & 0.16739\,{$\scriptstyle(\pm 0.00897)$} & \textbf{0.01496\,{$\scriptstyle(\pm 0.00214)$}} \\
WeightMix & 0.95162\,{$\scriptstyle(\pm 0.00309)$} & 0.22169\,{$\scriptstyle(\pm 0.01343)$} & 0.03144\,{$\scriptstyle(\pm 0.00225)$} \\
\rowcolor{oursrow}
WPME-logit & \textbf{0.95848\,{$\scriptstyle(\pm 0.00261)$}} & 0.18137\,{$\scriptstyle(\pm 0.00991)$} & 0.02555\,{$\scriptstyle(\pm 0.00227)$} \\
\rowcolor{oursrow}
WPME-prob & 0.95816\,{$\scriptstyle(\pm 0.00292)$} & \textbf{0.16484\,{$\scriptstyle(\pm 0.00905)$}} & 0.01499\,{$\scriptstyle(\pm 0.00209)$} \\
\bottomrule
\end{tabular}
\end{minipage}

\label{tab:ensemble}
\end{table*}

\paragraph{Validation protocol}
To avoid tuning mixing coefficients on the test set, we split the original test data into a validation split and a held-out test split. For each method, we select the hyperparameters ($\lambda, \beta$) that maximize validation-set accuracy, and then report accuracy, NLL, and ECE on the held-out test split.

In summary, WPME demonstrates that weight mixing is not merely an approximation of logit ensembling, but a complementary ingredient that can be leveraged for improved accuracy and calibration.

\section{Exploring the relationship between the predictions of the interpolated and the original models}

Here, we analyze the relationship between the BUMP model’s predictions and those of the original models.

\paragraph{Comparing the BUMP with the closer model}

For each model pair $(A,B)$, we count test datapoints where $M_1$’s top \#1 prediction matches $A$’s \#1, $A$’s \#2 prediction (second largest probability), or neither, and then average across the $50$ pairs. Results are shown in Table~\ref{tab:a12_hupli1_fixed_lambda}. In most cases $M_1$ agrees with $A$, when it does not, its prediction typically matches $A$'s second choice, while the remaining cases are negligible. This suggests that BUMPs primarily arise from small, local changes in the decision boundary: the interpolated model $M_1$ usually preserves $A$’s top-1 decision, and when it deviates it most often swaps to $A$’s runner-up rather than producing completely different predictions.

\begin{table}[ht!]
\centering
\caption{
    Comparing model $M_1$ with model $A$ on CIFAR-10, ResNet-20 with method Fixed-$\lambda$. Numbers are averages calculated for 50 model pairs.
}
\begin{tabular}{lccc}
\toprule
\multicolumn{4}{c}{Average number of datapoints} \\
\cmidrule(lr){1-4}
Width & A \#1 & A \#2 & Other \\
\midrule
4x-width  & $9909.72$ & $85.88$  & $4.40$  \\
8x-width  & $9893.92$ & $99.12$  & $6.96$  \\
16x-width & $9880.44$ & $110.80$ & $8.76$  \\
\bottomrule
\end{tabular}
\vspace{0.5cm}
\label{tab:a12_hupli1_fixed_lambda}
\end{table}

\paragraph{Comparing the BUMP with the farther model}
It is natural to assume that the accuracy gain of $M_1$ relative to $A$ reflects information contributed by the other model $B$. In this section, we test this hypothesis. We restrict to test datapoints, where the truthiness of the prediction (whether the predicted label matches the true label) \textit{differs} for the BUMP and the closer endpoint. 
We form two main categories: \emph{correct $\to$ incorrect} denotes those datapoints, where model $A$ predicts the true label and $M_1$ does not, and \emph{incorrect $\to$ correct} denotes those datapoints, where model $M_1$ predicts the true label and $A$ does not. (Note, that these datapoints characterize the accuracy gain of the BUMP.)
In each category, we count test datapoints where $M_1$’s top \#1 prediction matches $B$’s \#1, $B$’s \#2 prediction, or neither, and then average across the $50$ pairs. Thus, we quantify how the farther model influences the BUMP’s prediction, and whether it ``corrects'' or ``ruins'' the predictions of the closer model. For each model pair, we count datapoints in these overall six categories, and then average across the $50$ model pairs. As a control, we also introduce an independently chosen model $C$ for each pair to compare whether the BUMP’s predictions are more likely to match $B$, than those of $C$. Results are shown in Table~\ref{tab:hupli1_na_b12_tf_fixed_lambda}. Overall, we observe only minor differences between alignment with $B$ and alignment with $C$ in prediction. We therefore hypothesize that $B$ contributes by providing an ``informed'' regularization and sharpening $A$’s decision function through the interpolation direction, as opposed to causing $M_1$ to explicitly inherit $B$’s predicted labels. We leave the further exploration of this for future work.

\begin{table}[ht!]
\centering
\caption{Comparing the predictions of model $M_1$ with predictions from model $B$ and random chosen model $C$ on CIFAR-10, 16x-width ResNet-20 (50 model pairs) with method Fixed-$\lambda$. In each column, we present the average number of datapoints.}
\setlength{\tabcolsep}{6pt} 
\begin{tabular}{lccc ccc}
\toprule
\multicolumn{7}{c}{Average number of datapoints with changed truthiness} \\
\cmidrule(lr){1-7}
& \multicolumn{3}{c}{\emph{correct $\to$ incorrect}} & \multicolumn{3}{c}{\emph{incorrect $\to$ correct}} \\
\cmidrule(lr){2-4}\cmidrule(lr){5-7}
& \#1 & \#2 & Other & \#1 & \#2 & Other \\
\midrule
B & 10.30 & 18.94 & 11.02 & 40.72 & 13.78 & 5.92 \\
C & 10.80 & 18.18 & 11.28 & 39.26 & 15.04 & 6.12 \\
\bottomrule
\end{tabular}
\label{tab:hupli1_na_b12_tf_fixed_lambda}
\end{table}

\section{Conclusion}

We thoroughly analyzed BUMP, an accuracy peak near the endpoints of interpolation between permutation-aligned models that has not been explicitly discussed or analyzed in prior work. Leveraging BUMP, we proposed WPME, a simple hybrid of weight- and prediction-space mixing that consistently outperforms classical logit ensembling. Overall, our results demonstrate that aligned weight space offers practical opportunities for improved model combination.

\section*{Impact Statement}

This paper presents work whose goal is to advance the field of machine learning. There are many potential societal consequences of our work, none of which we feel must be specifically highlighted here.

\bibliographystyle{icml2026}
\bibliography{references}

\newpage
\appendix
\onecolumn

\section{Training details}
\label{appendix:training_details}

\paragraph{Tiny-10} Tiny-10 is a simple non-residual convolutional network where convolutional, normalization, and activation layers are following each other. Table~\ref{table:tiny-10} contains the precise architecture of the variant we were using in our experiments. We used Layer Normalization \citep{ba2016layer} to follow the methodology of \citet{ainsworth2022rebasin}. We trained our models for 300 epochs, we used SGD with momentum $0.9$. The initial learning rate was set to $0.01$, which was then divided by 10 at the third and the two thirds of the training. The batch size was $100$. We used weight decay with value $10^{-4}$. 

\begin{table}[h]
\caption{The Tiny-10 architecture we used.}
\label{table:tiny-10}
\centering
\begin{tabular}{l l}
\toprule
\multicolumn{2}{c}{Tiny-10}\\
\midrule
Layers & Name \\
\midrule
3 × 3 conv. 16-LN-ReLu & Layer 1 \\
3 × 3 conv. 16-LN-ReLu  & Layer 2 \\
3 × 3 conv. 32-LN-ReLu & Layer 3\\
3 × 3 conv. 32-LN-ReLu  & Layer 4 \\
3 × 3 conv. 32-LN-ReLu   & Layer 5 \\
3 × 3 conv. 64-LN-ReLu  & Layer 6 \\
3 × 3 conv. 64-LN-ReLu  & Layer 7 \\
1 × 1 conv. 64-LN-ReLu  & Layer 8 \\
Global average pooling  &  \\
Dense & \\
Logits & \\
\bottomrule
\end{tabular}
\end{table}

\paragraph{ResNet-20} We used the ResNet-20 architecture of \citet{resnet}, but with LayerNorms instead of BatchNorms. We trained our models for 250 epochs, we used SGD with momentum $0.9$. We applied a weight decay regularization term of $10^{-4}$, and used cosine decay schedule with linear warmup: the learning rate was initialized at $10^{-6}$ and linearly increased to $10^{-1}$ over an epoch and after a cosine decay schedule was used for the rest of the training. The batch size was $100$. 

\paragraph{Data augmentation}
The following data augmentation was performed during training for both models: 
\begin{itemize}[noitemsep]
    \item Random resizes of the image with scale factors between $0.8$ and $1.2$.
    \item Random $32\times 32$ pixel crops.
    \item Random horizontal flips.
    \item Random rotations between $\pm 30^\circ$.
\end{itemize}

\FloatBarrier
\section{Additional results for CIFAR-10 with ResNet-20 models}
\label{appendix:cifar10_moretables}

In this section, we present additional tables and figures to further support our experiments described in Sections~\ref{sec:robust}, \ref{sec:regularization}, \ref{sec:ensanalysis}, \ref{sec:novelensemble}. 

\subsection{Earlier checkpoints and pruning}

We report accuracy gains of the BUMPs in tables analogous to Table~\ref{tab:bumppermc10r20} and Table~\ref{tab:bumpnaivec10r20}. Results are shown for checkpoints at epoch $60$ (Table~\ref{tab:bumpc10r20ckpt60}), epoch $120$ (Table~\ref{tab:bumpc10r20ckpt120}) and epoch $180$ (Table~\ref{tab:bumpc10r20ckpt180}). 

Table~\ref{tab:bumpc10r20earlyaccs} summarizes the average accuracies across the $50$ model pairs ($A, B$) at epochs $60, 120, 180$ and $250$ (full training). 

We also examine pruning effects: BUMP accuracy gains when pruning is applied \textbf{before} alignment are given in Table~\ref{tab:bumpc10r20prunefirst} and gains when pruning is applied \textbf{after} alignment are shown in Table~\ref{tab:bumpc10r20prunelast}. For $16$x-width and $8$x-width we pruned $95\%$ of the weights, and for $4$x-width we pruned $90\%$ of the weights.
Finally, Table~\ref{tab:bumpc10r20prunedaccs} reports the average accuracies of the $50$ pruned model pairs. 

\begin{table}[h!]
\centering
\caption{
    Average accuracy gains of the BUMPs over their corresponding endpoint models on \textbf{aligned model pairs}. Results are reported on the CIFAR-10 test set using aligned ResNet-20 pairs (100 models) at \textbf{epoch 60}. Standard deviations are denoted in parentheses.
}
\begin{tabular}{lcc}
\toprule

\multicolumn{3}{c}{Accuracy gain \% --- Aligned pairs at epoch 60} \\
\cmidrule(lr){1-3}
Width & Fixed-$\lambda$ & Test \\
\midrule
4x-width  & $0.346\,{\scriptstyle(\pm 0.223)}$ & $0.663\,{\scriptstyle(\pm 0.528)}$ \\
8x-width  & $0.577\,{\scriptstyle(\pm 0.252)}$ & $1.009\,{\scriptstyle(\pm 0.549)}$ \\
16x-width & $0.731\,{\scriptstyle(\pm 0.309)}$ & $1.188\,{\scriptstyle(\pm 0.590)}$ \\
\bottomrule
\end{tabular}
\label{tab:bumpc10r20ckpt60}
\end{table}

\begin{table}[h!]
\centering
\caption{
    Average accuracy gains of the BUMPs over their corresponding endpoint models on \textbf{aligned model pairs}. Results are reported on the CIFAR-10 test set using aligned ResNet-20 pairs (100 models) at \textbf{epoch 120}. Standard deviations are denoted in parentheses.
}
\begin{tabular}{lcc}
\toprule
\multicolumn{3}{c}{Accuracy gain \% --- Aligned pairs at epoch 120} \\
\cmidrule(lr){1-3}
Width & Fixed-$\lambda$ & Test \\
\midrule
4x-width  & $0.291\,{\scriptstyle(\pm 0.178)}$ & $0.506\,{\scriptstyle(\pm 0.365)}$ \\
8x-width  & $0.448\,{\scriptstyle(\pm 0.191)}$ & $0.732\,{\scriptstyle(\pm 0.380)}$ \\
16x-width & $0.614\,{\scriptstyle(\pm 0.261)}$ & $0.970\,{\scriptstyle(\pm 0.487)}$ \\
\bottomrule
\end{tabular}
\label{tab:bumpc10r20ckpt120}
\end{table}

\begin{table}[h!]
\centering
\caption{
    Average accuracy gains of the BUMPs over their corresponding endpoint models on \textbf{aligned model pairs}. Results are reported on the CIFAR-10 test set using aligned ResNet-20 pairs (100 models) at \textbf{epoch 180}. Standard deviations are denoted in parentheses.
}
\begin{tabular}{lcc}
\toprule
\multicolumn{3}{c}{Accuracy gain \% --- Aligned pairs at epoch 180} \\
\cmidrule(lr){1-3}
Width & Fixed-$\lambda$ & Test \\
\midrule
4x-width  & $0.223\,{\scriptstyle(\pm 0.148)}$ & $0.354\,{\scriptstyle(\pm 0.250)}$ \\
8x-width  & $0.358\,{\scriptstyle(\pm 0.173)}$ & $0.526\,{\scriptstyle(\pm 0.300)}$ \\
16x-width & $0.513\,{\scriptstyle(\pm 0.187)}$ & $0.755\,{\scriptstyle(\pm 0.348)}$ \\
\bottomrule
\end{tabular}
\label{tab:bumpc10r20ckpt180}
\end{table}

\begin{table}[h!]
\centering
\caption{
    Average model accuracy for CIFAR-10, ResNet-20 at different training epochs, comparing Model A and Model B for various model widths (100 models).
}
\begin{tabular}{lcccccccc}
\toprule
& \multicolumn{8}{c}{Average Accuracy} \\
\cmidrule(lr){2-9}
\addlinespace
Width & \multicolumn{2}{c}{At epoch 60} & \multicolumn{2}{c}{At epoch 120} & \multicolumn{2}{c}{At epoch 180} & \multicolumn{2}{c}{At epoch 250} \\
& A & B & A & B & A & B & A & B \\
\midrule
4x-width  & 0.8729 & 0.8736 & 0.8966 & 0.8945 & 0.9126 & 0.9138 & 0.9351 & 0.9356 \\
8x-width  & 0.8869 & 0.8858 & 0.9081 & 0.9069 & 0.9206 & 0.9196 & 0.9418 & 0.9412 \\
16x-width & 0.8980 & 0.8935 & 0.9135 & 0.9131 & 0.9251 & 0.9249 & 0.9450 & 0.9456 \\
\bottomrule
\end{tabular}
\vspace{0.5cm}
\label{tab:bumpc10r20earlyaccs}
\end{table}

\begin{table}[h!]
\centering

\begin{minipage}[t]{0.48\textwidth}
\centering
\caption{
    Average accuracy gains of the BUMPs over their corresponding endpoint models when pruning is applied \textbf{before} alignment. Results are reported on the CIFAR-10 test set using ResNet-20 pairs (100 models) trained to convergence (250 epochs). Standard deviations are denoted in parentheses.
}
\begin{tabular}{lcc}
\toprule
\multicolumn{3}{c}{Accuracy gain \% --- Pruned before alignment} \\
\cmidrule(lr){1-3}
Width & Fixed-$\lambda$ & Test \\
\midrule
4x-width  & $1.592\,{\scriptstyle(\pm 0.946)}$ & $3.317\,{\scriptstyle(\pm 2.705)}$ \\
8x-width  & $2.359\,{\scriptstyle(\pm 1.640)}$ & $6.071\,{\scriptstyle(\pm 4.958)}$ \\
16x-width & $1.536\,{\scriptstyle(\pm 1.072)}$ & $2.171\,{\scriptstyle(\pm 1.819)}$ \\
\bottomrule
\end{tabular}
\label{tab:bumpc10r20prunefirst}
\end{minipage}
\hfill
\begin{minipage}[t]{0.48\textwidth}
\centering
\caption{
    Average accuracy gains of the BUMPs over their corresponding endpoint models when pruning is applied \textbf{after} alignment. Results are reported on the CIFAR-10 test set using ResNet-20 pairs (100 models) trained to convergence (250 epochs). Standard deviations are denoted in parentheses.
}
\begin{tabular}{lcc}
\toprule
\multicolumn{3}{c}{Accuracy gain \% --- Pruned after alignment} \\
\cmidrule(lr){1-3}
Width & Fixed-$\lambda$ & Test \\
\midrule
4x-width  & $1.933\,{\scriptstyle(\pm 1.057)}$ & $4.775\,{\scriptstyle(\pm 3.487)}$ \\
8x-width  & $2.296\,{\scriptstyle(\pm 1.635)}$ & $6.300\,{\scriptstyle(\pm 5.133)}$ \\
16x-width & $1.603\,{\scriptstyle(\pm 1.108)}$ & $2.550\,{\scriptstyle(\pm 2.066)}$ \\
\bottomrule
\end{tabular}
\label{tab:bumpc10r20prunelast}
\end{minipage}
\end{table}

\begin{table}[h!]
\centering
\caption{
    Average model accuracies for CIFAR-10, ResNet-20 across widths, comparing original and pruned variants of Model A and Model B (100 models).
}
\begin{tabular}{lcccc}
\toprule
& \multicolumn{4}{c}{Average Accuracies} \\
\cmidrule(lr){2-5}
Width & Model A & Model B & Pruned Model A & Pruned Model B \\
\midrule
4x-width  & 0.9351 & 0.9356 & 0.6195 & 0.6235 \\
8x-width  & 0.9418 & 0.9412 & 0.4613 & 0.4652 \\
16x-width & 0.9450 & 0.9456 & 0.7433 & 0.7493 \\
\bottomrule
\end{tabular}
\vspace{0.5cm}
\label{tab:bumpc10r20prunedaccs}
\end{table}

\subsection{Noise models}

We present the results for our homoscedastic and heteroscedastic noise model experiments in Table~\ref{tab:noise_model_gains}. 

\begin{table}[t]
\centering
\caption{Accuracy gains (mean, std, $p$-value) for homoskedastic and heteroskedastic noise models across widths.}
\begin{tabular}{ccccccc}
\toprule
& \multicolumn{3}{c}{Homoskedastic} & \multicolumn{3}{c}{Heteroskedastic} \\
\cmidrule(lr){2-4} \cmidrule(lr){5-7}
Width & mean & std & $p$ & mean & std & $p$ \\
\midrule
4x-width & -0.0001 & 0.0006 & 0.146 & -0.0005 & 0.0007 & 1.27e-05 \\
8x-width & -0.0001 & 0.0007 & 0.22 & -0.0007 & 0.0009 & 7.4e-07 \\
16x-width & -0.0002 & 0.0006 & 0.0186 & -0.0009 & 0.0010 & 2.27e-07 \\
\bottomrule
\end{tabular}
\label{tab:noise_model_gains}
\end{table}

\subsection{Further figures to explore the Weight-Prediction Mixed Ensembling}

\begin{figure}[!htbp]
\centering
\includegraphics[width=0.48\linewidth]{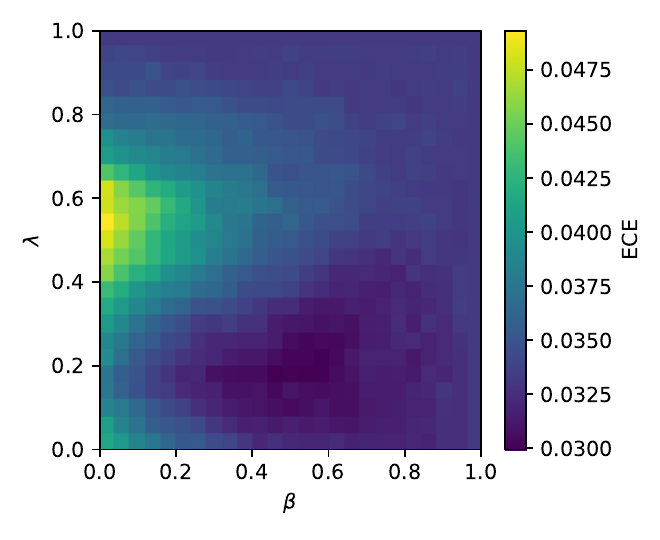}
\includegraphics[width=0.48\linewidth]{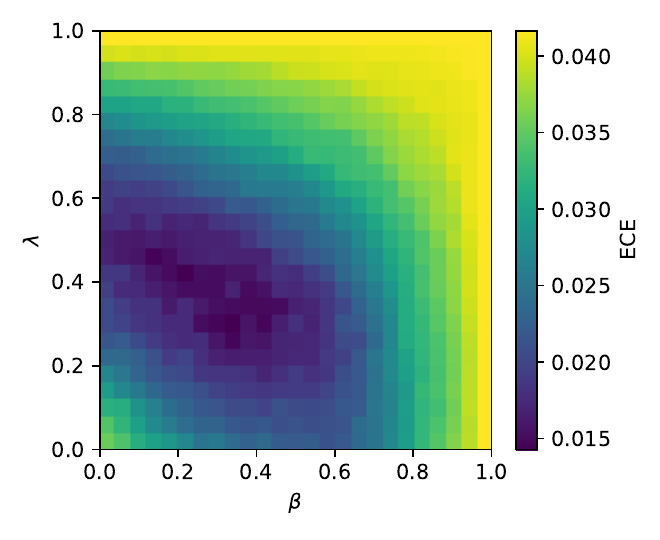}
\caption{
Expected calibration error (ECE; 15 bins) over $(\lambda,\beta)$ for probability-space hybrid ensembling,
where weight-interpolated models are combined with endpoint predictions after softmax.
Lower values indicate improved calibration.
}
\label{fig:ensemble_ece_surface}
\end{figure}

\FloatBarrier
\section{Results for CIFAR-100 with ResNet-20 models}
\label{appendix:cifar100}

In this section, we present the figures and tables for experiments on the CIFAR-100 dataset with ResNet-20 models. For each width multiplier in $\{4, 8, 16\}$ we train $50$ models, form $25$ model pairs, and align them via the weight-matching algorithm of \citet{ainsworth2022rebasin}. We used a similar methodology to quantify accuracy gain as in Section~\ref{sec:bumpdemo}, but with $\lambda = 0.04$ for $4$x-width, $\lambda = 0.07$ for $8$x-width, $\lambda = 0.08$ for $16$x-width with the Fixed-$\lambda$ method.

\subsection{The BUMPs for aligned and naïve model pairs}

We present figures analogous to Figures~\ref{fig:bumpwm16} and \ref{fig:bumpnaivewm16}, together with tables analogous to Tables~\ref{tab:bumppermc10r20}, and \ref{tab:bumpnaivec10r20}. 

\begin{figure}
\centering
    \subfloat[Left BUMP (aligned)]{%
            \includegraphics[trim=0cm 0cm 0cm 0cm, clip=true, width=0.5\columnwidth]{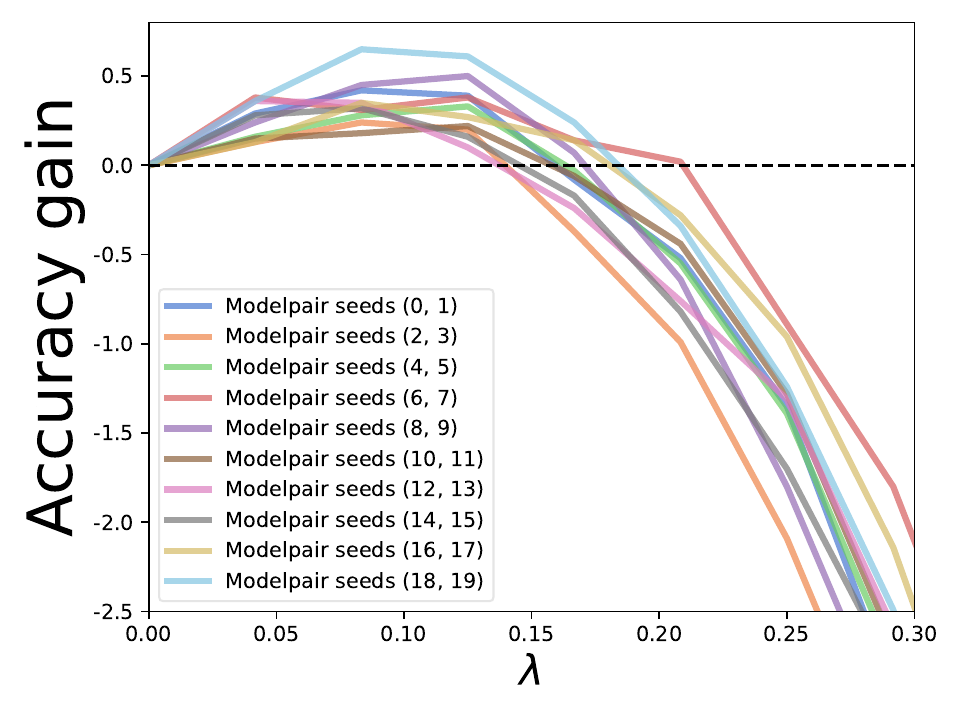}%
        \label{fig:bumpleftC100}%
        }%
    \subfloat[Right BUMP (aligned)]{%
        \includegraphics[trim=0cm 0cm 0cm 0cm, clip=true, width=0.5\columnwidth]{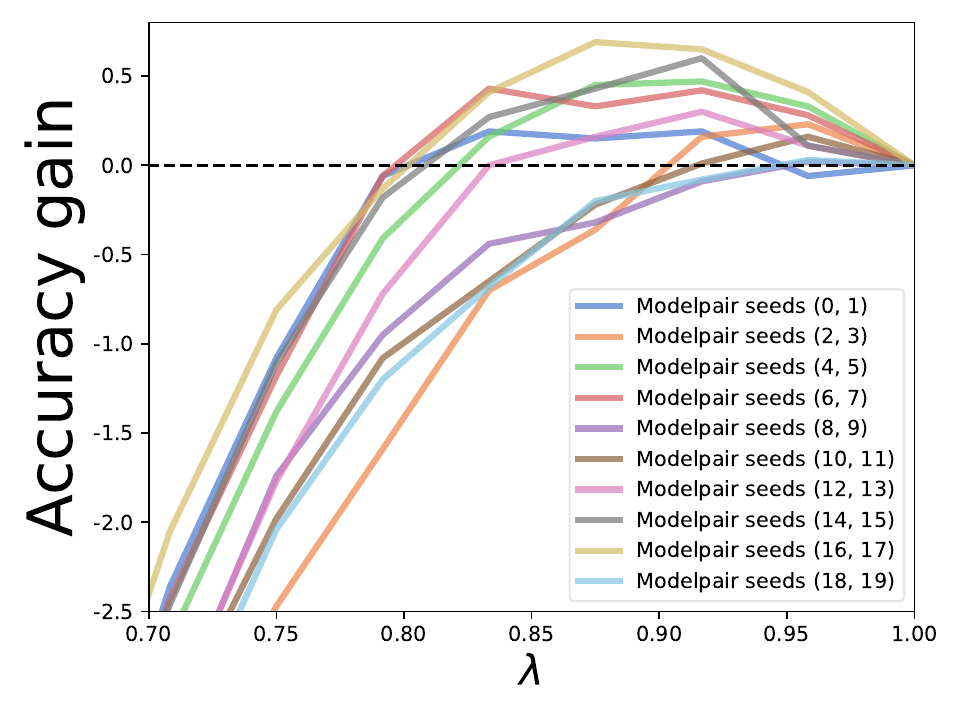}%
        \label{fig:bumprightC100}%
        }%
    \caption{The two BUMPs when interpolating between aligned models, the accuracies are measured on the CIFAR-100 test set, models are ResNet-20, showing $10$ model pairs. Width multiplier is 16. We shifted the plots so that all accuracies of the original models are 0.}%
    \label{fig:bumpwm16C100}%
\end{figure}

\begin{figure}
\centering
    \subfloat[Left BUMP (na\"ive)]{%
        \includegraphics[trim=0cm 0cm 0cm 0cm, clip=true, width=0.5\columnwidth]{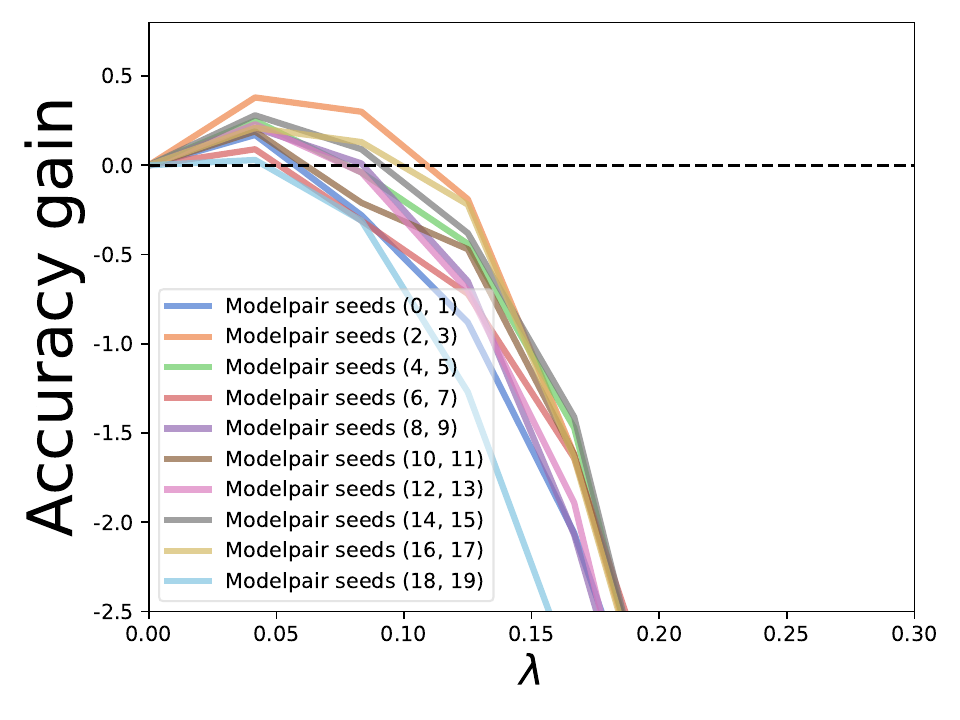}%
        \label{fig:bumpnaiveleftC100}%
        }%
    \subfloat[Right BUMP (na\"ive)]{%
        \includegraphics[trim=0cm 0cm 0cm 0cm, clip=true, width=0.5\columnwidth]{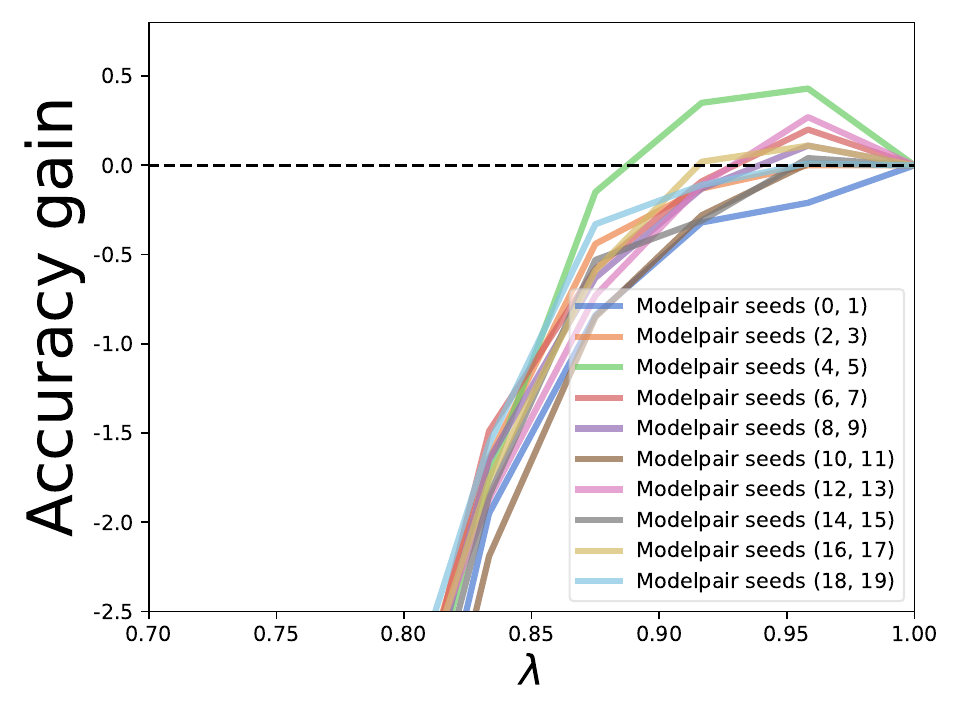}%
        \label{fig:bumpnaiverightC100}%
        }%
    \caption{The two BUMPs when interpolating between models without alignment, the accuracies are measured on the CIFAR-100 test set, models are ResNet-20, showing $10$ model pairs. Width multiplier is 16. We shifted the plots so that all accuracies of the original models are 0.}%
    \label{fig:bumpnaiveC100}%
\end{figure}

For the $16$x-width setting, we plot interpolation curves for $\lambda \in (0.0, 0.3)$ and $\lambda \in(0.7, 1.0)$ across $10$ model pairs. Figure~\ref{fig:bumpwm16C100} shows clear BUMPs near both endpoints for aligned model pairs, whereas for naïve pairs the BUMP is absent or much smaller (Figure~\ref{fig:bumpnaiveC100}). We report results for both methods quantifying accuracy gain of a BUMP in Table~\ref{tab:bumpperm_C100R20} (aligned pairs) and Table~\ref{tab:bumpnaive_C100R20} (naïve pairs). 

\begin{table}[h!]
\centering
\begin{minipage}[t]{0.48\textwidth}
\centering
\caption{
    Average accuracy gains of the BUMPs over their corresponding endpoint models on \textbf{aligned model pairs}. Results are reported on the CIFAR-100 test set using aligned ResNet-20 pairs (50 models) trained to convergence (250 epochs). Standard deviations are denoted in parentheses.
}
\begin{tabular}{lcc}
\toprule
\multicolumn{3}{c}{Accuracy gain \% --- Aligned pairs} \\
\cmidrule(lr){1-3}
Width & Fixed-$\lambda$ & Test \\
\midrule
4x-width  & $0.201\,{\scriptstyle(\pm 0.197)}$ & $0.254\,{\scriptstyle(\pm 0.218)}$ \\
8x-width  & $0.226\,{\scriptstyle(\pm 0.184)}$ & $0.264\,{\scriptstyle(\pm 0.169)}$ \\
16x-width & $0.319\,{\scriptstyle(\pm 0.188)}$ & $0.366\,{\scriptstyle(\pm 0.173)}$ \\
\bottomrule
\end{tabular}
\label{tab:bumpperm_C100R20}
\end{minipage}\hfill
\begin{minipage}[t]{0.48\textwidth}
\centering
\caption{
    Average accuracy gains of the BUMPs over their corresponding endpoint models on \emph{non-aligned}, i.e., \textbf{naïve model pairs}. Results are reported on the CIFAR-100 test set using aligned ResNet-20 pairs (50 models) trained to convergence (250 epochs). Standard deviations are denoted in parentheses.
}
\begin{tabular}{lcc}
\toprule
\multicolumn{3}{c}{Accuracy gain \% --- Naïve pairs} \\
\cmidrule(lr){1-3}
Width & Fixed-$\lambda$ & Test \\
\midrule
4x-width  & $0.092\,{\scriptstyle(\pm 0.289)}$  & $0.156\,{\scriptstyle(\pm 0.203)}$ \\
8x-width  & $-0.037\,{\scriptstyle(\pm 0.290)}$ & $0.140\,{\scriptstyle(\pm 0.154)}$ \\
16x-width & $-0.007\,{\scriptstyle(\pm 0.218)}$ & $0.178\,{\scriptstyle(\pm 0.142)}$ \\
\bottomrule
\end{tabular}
\label{tab:bumpnaive_C100R20}
\end{minipage}
\end{table}

\subsection{A geometric and probabilistic analysis of model mixing}

\begin{figure}[htbp]
  \centering
  \includegraphics[width=0.8\textwidth]{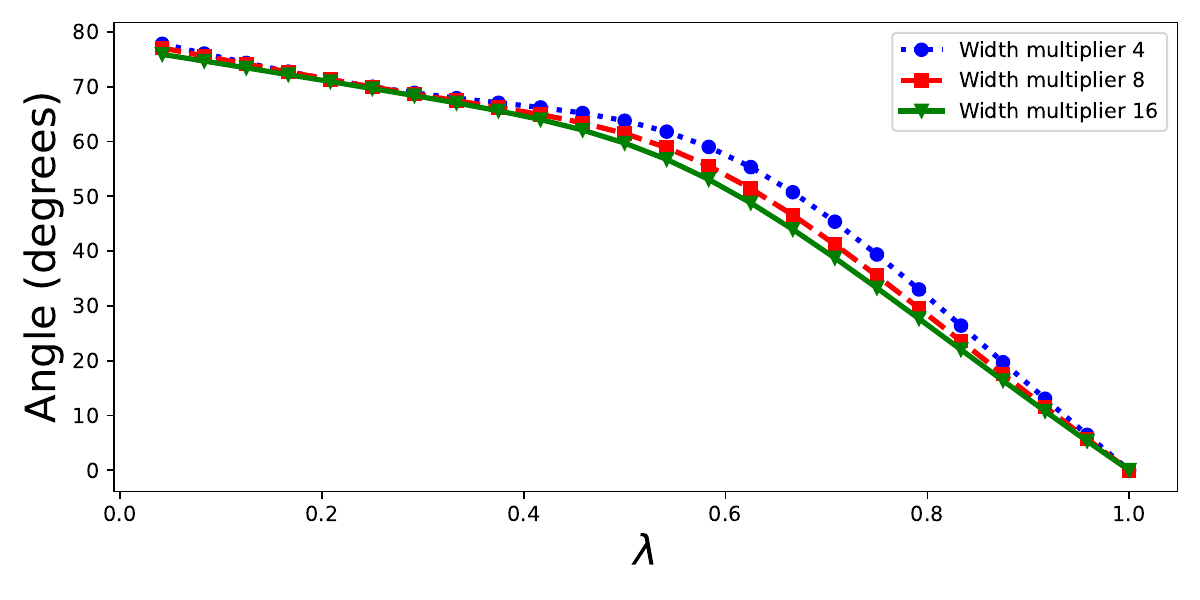}  
  \caption{The mean angles across the test datapoints for $g = 24$, CIFAR-100, ResNet-20 ($50$ models).}
  \label{fig:meanangles_cifar100}
\end{figure}

\begin{figure}[tbp]
  \subfloat[Accuracy curves]{%
    \centering
    \includegraphics[width=0.48\columnwidth]{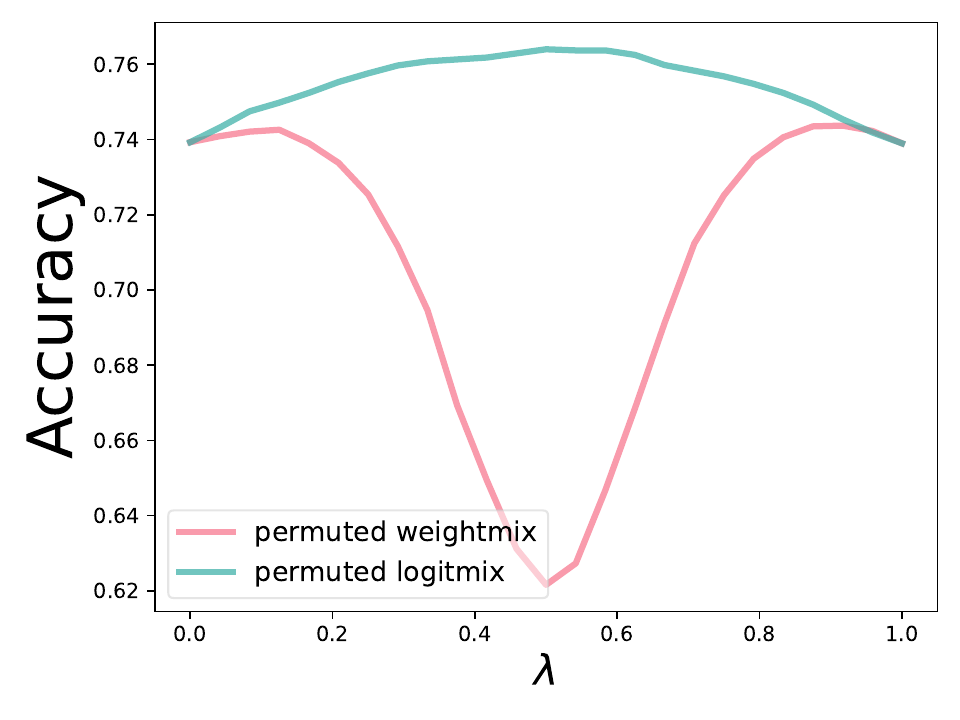}
      \label{fig:logitweightC100}
  }%
  \subfloat[KL divergences]{
    \includegraphics[width=0.48\columnwidth]{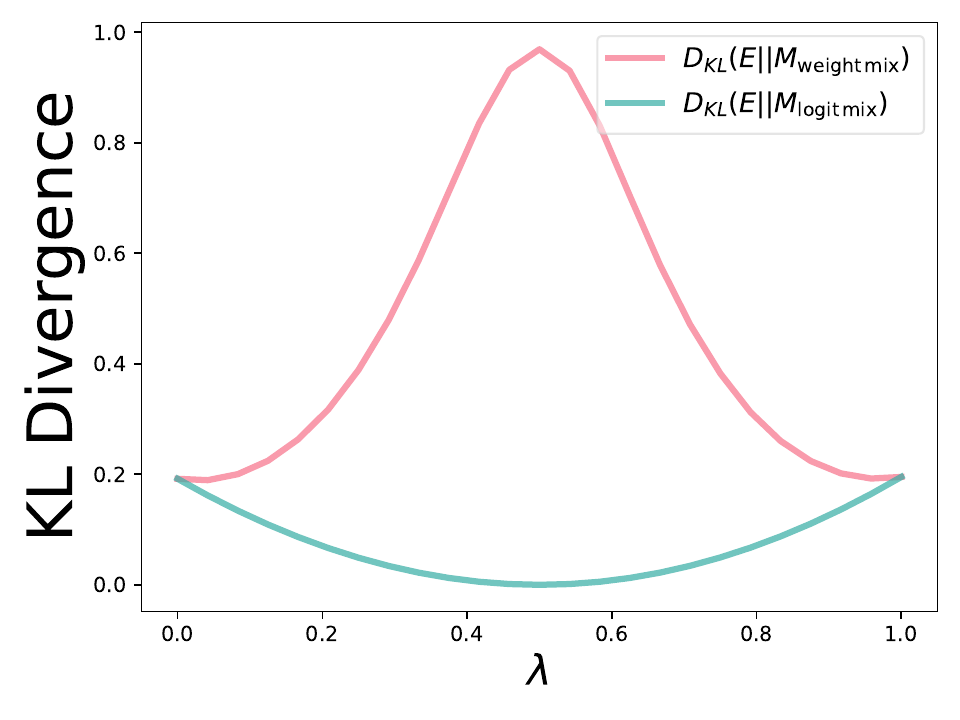}
      \label{fig:kl_div_wmlmEC100}
  }%
  \caption{\textbf{Left (comparing accuracy curves):} The WeightMix and LogitMix accuracy curve on CIFAR-100, ResNet-20, width multiplier is $16$. \textbf{Right (comparing class probabilities):} Kullback-Leibler divergence between the class probability distributions of the WeightMix/LogitMix models and the $\lambda=0.5$ LogitMix (classicle ensemble) model, CIFAR-100, ResNet-20.}
\end{figure}

In this section, we present the results for the experiments described in Section~\ref{sec:ensanalysis} for CIFAR-100 and ResNet-20. Figure~\ref{fig:meanangles_cifar100} depicts the mean angles between the WeightMix and LogitMix logits. Figure~\ref{fig:logitweightC100} shows the accuracy curves, and Figure~\ref{fig:kl_div_wmlmEC100} shows the KL-divergences between the Weightmix and LogitMix models. 

\FloatBarrier
\section{Results for CIFAR-10 with Tiny-10 models}
\label{appendix:tiny10}

In this section, we present the figures and tables for experiments on the CIFAR-10 dataset with Tiny-10 models. For each width multiplier in $\{4, 8, 16\}$ we train $50$ models, form $25$ model pairs, and align them via the weight-matching algorithm of \citet{ainsworth2022rebasin}. We used a similary methodology to quantify accuracy gain as in Section~\ref{sec:bumpdemo}, but with $\lambda = 0.003$ for $4$x-width, $\lambda = 0.01$ for $8$x-width, $\lambda = 0.015$ for $16$x-width in the Fixed-$\lambda$ method. 

\subsection{The BUMPs for aligned and naïve model pairs}

We present figures analogous to Figures~\ref{fig:bumpwm16} and \ref{fig:bumpnaivewm16}, together with tables analogous to Tables~\ref{tab:bumppermc10r20}, and \ref{tab:bumpnaivec10r20}. 

\begin{figure}
\centering
    \subfloat[Left BUMP (aligned)]{%
            \includegraphics[trim=0cm 0cm 0cm 0cm, clip=true, width=0.5\columnwidth]{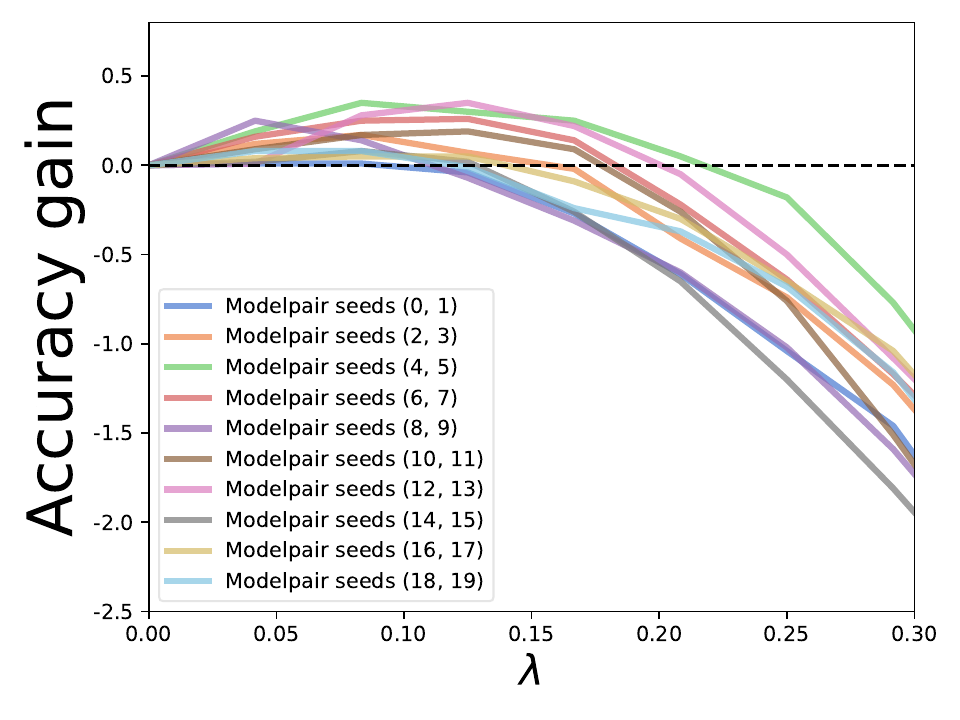}%
        \label{fig:bumpleftTiny}%
        }%
    \subfloat[Right BUMP (aligned)]{%
        \includegraphics[trim=0cm 0cm 0cm 0cm, clip=true, width=0.5\columnwidth]{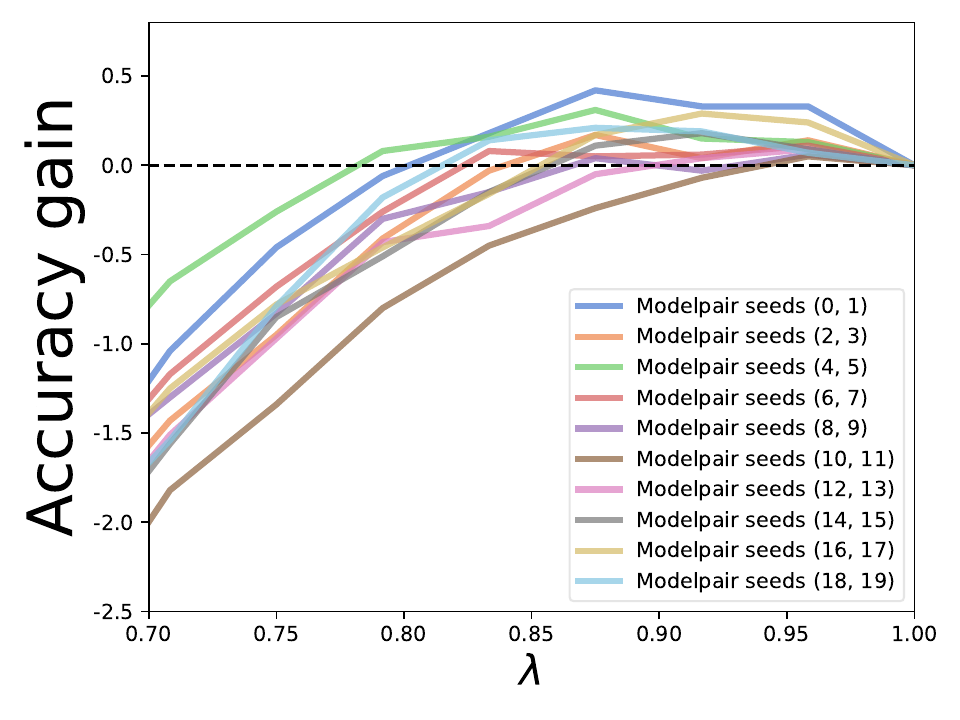}%
        \label{fig:bumprightTiny}%
        }%
    \caption{The two BUMPs when interpolating between aligned models, the accuracies are measured on the CIFAR-10 test set, models are Tiny-10, showing $10$ model pairs. Width multiplier is 16. We shifted the plots so that all accuracies of the original models are 0.}%
    \label{fig:bumpwm16Tiny}%
\end{figure}

\begin{figure}
\centering
    \subfloat[Left BUMP (na\"ive)]{%
        \includegraphics[trim=0cm 0cm 0cm 0cm, clip=true, width=0.5\columnwidth]{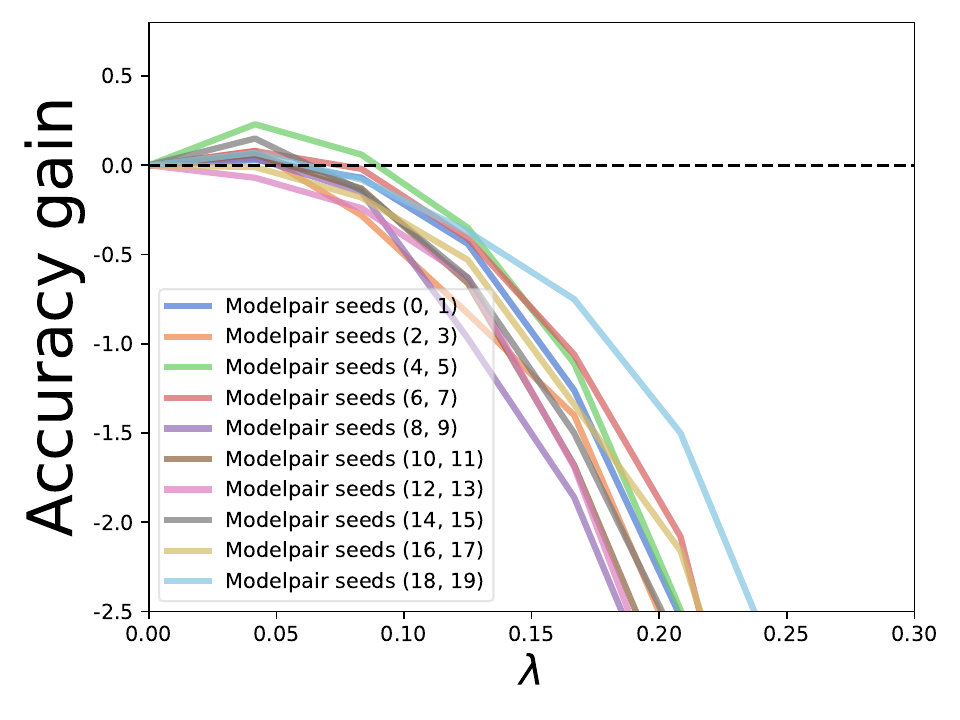}%
        \label{fig:bumpnaiveleftTiny}%
        }%
    \subfloat[Right BUMP (na\"ive)]{%
        \includegraphics[trim=0cm 0cm 0cm 0cm, clip=true, width=0.5\columnwidth]{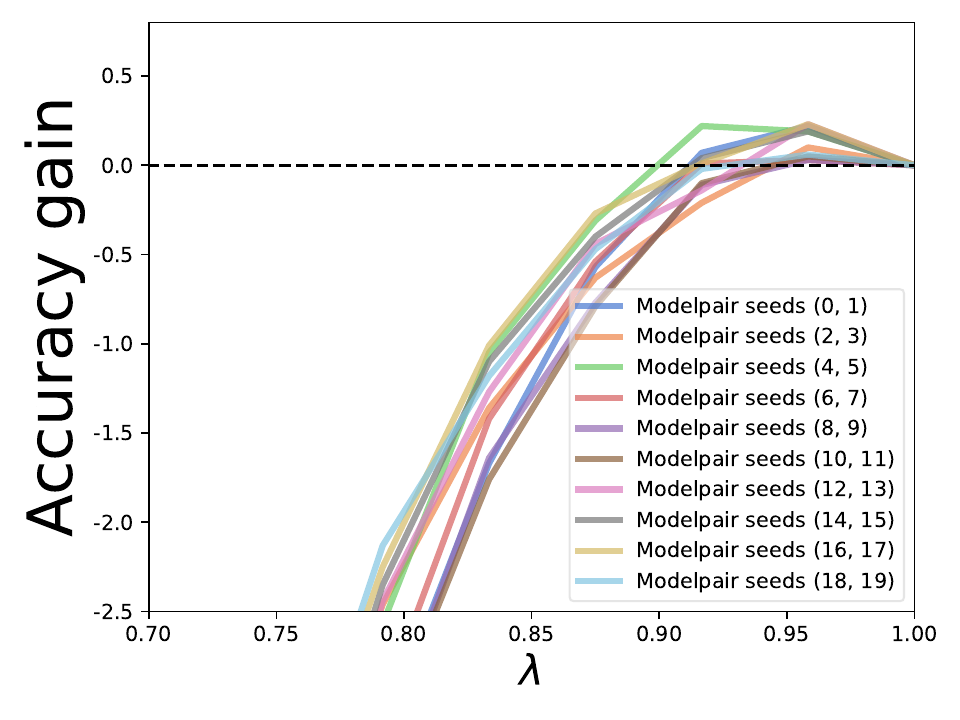}%
        \label{fig:bumpnaiverightTiny}%
        }%
    \caption{The two BUMPs when interpolating between models without alignment, the accuracies are measured on the CIFAR-10 test set, models are Tiny-10, showing $10$ model pairs. Width multiplier is 16. We shifted the plots so that all accuracies of the original models are 0.}%
    \label{fig:bumpnaiveTiny}%
\end{figure}

For the $16$x-width setting, we plot interpolation curves for $\lambda \in (0.0, 0.3)$ and $\lambda \in(0.7, 1.0)$ across $10$ model pairs. Figure~\ref{fig:bumpwm16Tiny} shows clear BUMPs near both endpoints for aligned model pairs, whereas for naïve pairs the BUMP is absent or much smaller (Figure~\ref{fig:bumpnaiveTiny}). We report results for both methods quantifying accuracy gain of a BUMP in Table~\ref{tab:hupli_tiny10_perm} (aligned pairs) and Table~\ref{tab:hupli_tiny10_naive} (naïve pairs).

\begin{table}[h!]
\centering
\begin{minipage}[t]{0.48\textwidth}
\centering
\caption{
    Average accuracy gains of the BUMPs over their corresponding endpoint models on \textbf{aligned model pairs}. Results are reported on the CIFAR-10 test set using aligned Tiny-10 pairs (50 models) trained to convergence (300 epochs). Standard deviations are denoted in parentheses.
}
\begin{tabular}{lcc}
\toprule
\multicolumn{3}{c}{Accuracy gain \% --- Aligned pairs} \\
\cmidrule(lr){1-3}
Width & Fixed-$\lambda$ & Test \\
\midrule
4x-width  & $0.013\,{\scriptstyle(\pm 0.024)}$ & $0.106\,{\scriptstyle(\pm 0.122)}$ \\
8x-width  & $0.040\,{\scriptstyle(\pm 0.054)}$ & $0.144\,{\scriptstyle(\pm 0.131)}$ \\
16x-width & $0.054\,{\scriptstyle(\pm 0.056)}$ & $0.178\,{\scriptstyle(\pm 0.115)}$ \\
\bottomrule
\end{tabular}
\label{tab:hupli_tiny10_perm}
\end{minipage}\hfill
\begin{minipage}[t]{0.48\textwidth}
\centering
\caption{
    Average accuracy gains of the BUMPs over their corresponding endpoint models on \emph{non-aligned}, i.e., \textbf{naïve model pairs}. Results are reported on the CIFAR-10 test set using aligned Tiny-10 pairs (50 models) trained to convergence (300 epochs). Standard deviations are denoted in parentheses.
}
\begin{tabular}{lcc}
\toprule
\multicolumn{3}{c}{Accuracy gain \% --- Naïve pairs} \\
\cmidrule(lr){1-3}
Width & Fixed-$\lambda$ & Test \\
\midrule
4x-width  & $0.027\,{\scriptstyle(\pm 0.038)}$ & $0.077\,{\scriptstyle(\pm 0.112)}$ \\
8x-width  & $0.075\,{\scriptstyle(\pm 0.087)}$ & $0.112\,{\scriptstyle(\pm 0.133)}$ \\
16x-width & $0.050\,{\scriptstyle(\pm 0.059)}$ & $0.093\,{\scriptstyle(\pm 0.081)}$ \\
\bottomrule
\end{tabular}
\label{tab:hupli_tiny10_naive}
\end{minipage}

\end{table}

\subsection{A geometric and probabilistic analysis of model mixing}

\begin{figure}[htbp]
  \centering
  \includegraphics[width=0.8\textwidth]{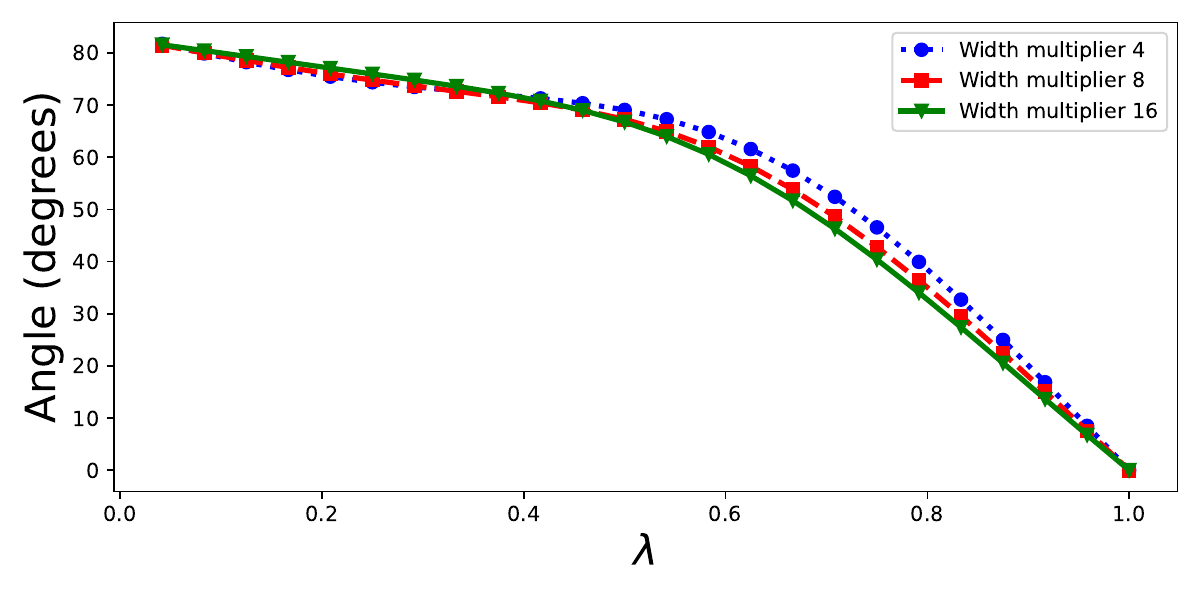}  
  \caption{The mean angles across the test datapoints for $g = 24$, measured on CIFAR-10, model is Tiny-10 ($50$ models).}
  \label{fig:meanangles_tiny10}
\end{figure}

\begin{figure}[tbp]
  \subfloat[Accuracy curves]{%
    \centering
    \includegraphics[width=0.48\columnwidth]{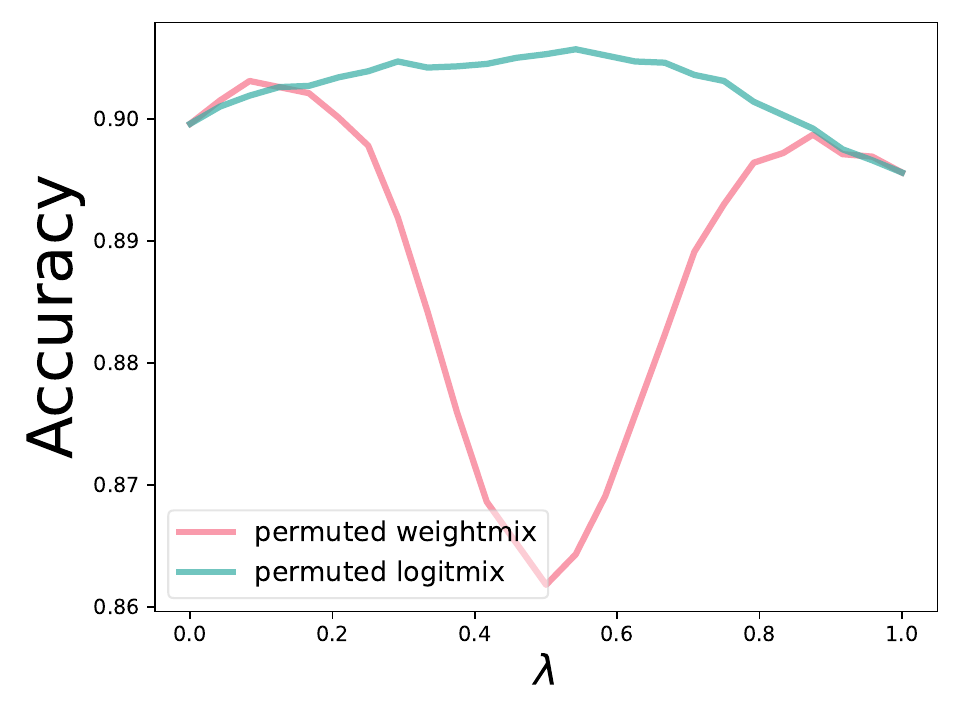}
      \label{fig:logitweightTiny}
  }%
  \subfloat[KL divergences]{
    \includegraphics[width=0.48\columnwidth]{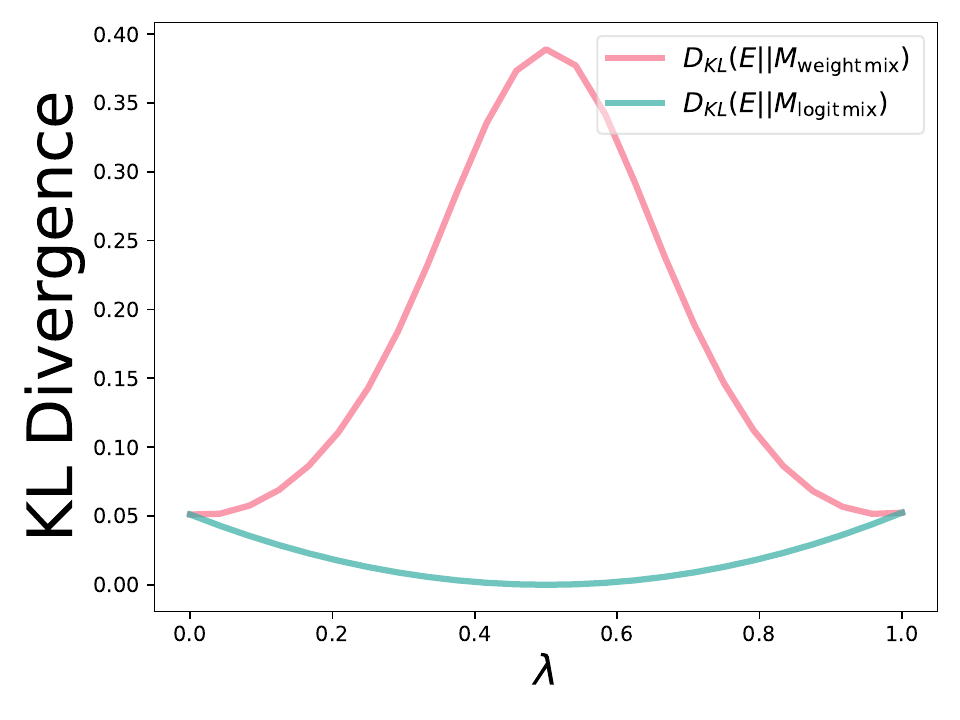}
      \label{fig:kl_div_wmlmETiny}
  }%
  \caption{\textbf{Left (comparing accuracy curves):} The WeightMix and LogitMix accuracy curve on CIFAR-10 with model Tiny-10, width multiplier is $16$. \textbf{Right (comparing class probabilities):} Kullback-Leibler divergence between the class probability distributions of the WeightMix/LogitMix models and the $\lambda=0.5$ LogitMix (classical logit ensemble) model, dataset is CIFAR-10, model is Tiny-10, width multiplier is $16$.}
\end{figure}

In this section, we present the results for the experiments described in Section~\ref{sec:ensanalysis} for CIFAR-10 and Tiny-10. Figure~\ref{fig:meanangles_tiny10} depicts the mean angles between the WeightMix and LogitMix logits. Figure~\ref{fig:logitweightTiny} shows the accuracy curves, and Figure~\ref{fig:kl_div_wmlmETiny} shows the KL-divergences between the Weightmix and LogitMix models. 

\end{document}